%% file: main.tex
\documentclass{article}

\usepackage{arxiv}
\usepackage[utf8]{inputenc} 
\usepackage[T1]{fontenc}    
\usepackage{hyperref}       
\usepackage{url}            
\usepackage{booktabs}       
\usepackage{amsfonts}       
\usepackage{nicefrac}       
\usepackage{microtype}      
\usepackage{subfig}
\usepackage{lipsum}
\usepackage{caption}
\usepackage{graphics}
\usepackage{graphicx}
\usepackage{algorithm2e}
\usepackage{cite}
\usepackage{comment}
\usepackage{float}
\usepackage{amsmath}
\usepackage{multirow}
\usepackage[dvipsnames]{xcolor}
\usepackage[title]{appendix}
\usepackage{ulem}
\usepackage{cancel}
\usepackage{tabularx}
\usepackage{colortbl}

\usepackage{xparse}
\NewDocumentCommand{\scnum}{ >{\SplitArgument{1}{e}}m }
 {\scnumaux#1}
\NewDocumentCommand{\scnumaux}{ m m }
 {#1\,\mathrm{e}{#2}}

\usepackage{lineno}
\usepackage[onehalfspacing]{setspace}

\usepackage{amsmath}

\DeclareMathOperator*{\argmax}{\arg\!\max}
\DeclareMathOperator*{\argmin}{\arg\!\min}
\DeclareMathOperator*{\MMD}{MMD}
\DeclareMathOperator*{\PRD}{PRD}

\DeclareMathOperator*{\Dis}{DPR}
\DeclareMathOperator*{\Dip}{DiP}
\DeclareMathOperator*{\Dir}{DiR}
\DeclareMathOperator*{\NN}{NN}
\DeclareMathOperator*{\proj}{proj}

\title{Data-efficient visuomotor policy training using reinforcement learning and generative models}

\author{
  \thanks{denotes an equal contribution} Ali Ghadirzadeh \thanks{KTH Royal Institute of Technology, Division of Robotics, Perception and Learning} \, \thanks{Aalto University, Intelligent Robotics Research Group}\\
  \texttt{algh@kth.se} \\
  \And
  $^*$Petra Poklukar,$^{\dagger}$ \\
  \texttt{poklukar@kth.se} \\
  \AND
  Ville Kyrki\,$^{\ddagger}$ \\
  \texttt{ville.kyrki@aalto.fi} \\ 
  \And
  Danica Kragic\,$^{\dagger}$ \\
  \texttt{dani@kth.se} \\ 
  \And
  M\aa rten Bj\"orkman\,$^{\dagger}$ \\
  \texttt{celle@kth.se} \\ 
}

\begin{document}
\maketitle

\input{inputs/abstract.tex}


\input{inputs/introduction.tex}

\input{inputs/related_work.tex}

\input{inputs/background.tex}
\input{inputs/policy_training.tex}
\input{inputs/generative_model_training.tex}
\input{inputs/experiments.tex}

\input{inputs/conclusion.tex}

\section*{Acknowledgments}
This work was supported by Knut and Alice Wallenberg Foundation, the EU through the project EnTimeMent, the Swedish Foundation for Strategic Research through the COIN project, and also by the  Academy of Finland through the DEEPEN project.

\bibliographystyle{IEEEtran}
\bibliography{ref}

\newpage
\appendix
\input{inputs/appendix}

\end{document}

%% file: inputs/abstract.tex
\begin{abstract}
We present a data-efficient framework for solving visuomotor sequential decision-making problems which exploits the combination of reinforcement learning (RL) and latent variable generative models. Our framework trains deep visuomotor policies by introducing an action latent variable such that the feed-forward policy search can be divided into three parts: (i) training a sub-policy that outputs a distribution over the action latent variable given a state of the system, (ii) unsupervised training of a generative model that outputs a sequence of motor actions conditioned on the latent action variable, and (iii) supervised training of the deep visuomotor policy in an end-to-end fashion. Our approach enables safe exploration and alleviates the data-inefficiency problem as it exploits prior knowledge about valid sequences of motor actions. Moreover, we provide a set of measures for evaluation of generative models such that we are able to predict the performance of the RL policy training prior to the actual training on a physical robot. We define two novel measures of disentanglement and local linearity for assessing the quality of latent representations, and complement them with existing measures for assessment of the learned distribution. We experimentally determine the characteristics of different generative models that have the most influence on performance of the final policy training on a robotic picking task.
\end{abstract}

%% file: inputs/introduction.tex
\section{Introduction}
\label{sec:introduction}

Reinforcement learning (RL) can leverage modeling capability of generative models to solve complex sequential decision making problems more efficiently \cite{ghadirzadeh2017deep, arndt2019meta}. 
RL has been applied to end-to-end training of deep visuomotor robotic policies \cite{levine2016end,levine2018learning}
but it is typically too data-inefficient 
especially when applied to tasks that provide only a terminal reward at the end of an episode. 
One way to alleviate the data-inefficiency problem in RL is by leveraging prior knowledge to reduce the complexity of the optimization problem.
One prior that significantly reduces the data requirement is an approximation of the distribution from which valid action sequences can be sampled.
Such distributions can be efficiently approximated by training generative models given a sufficient amount of valid action sequences. 

The question is then how to combine powerful RL optimization algorithms with the modeling capability of generative models to improve the efficiency of the policy training?
Moreover, which characteristics of the generative models are important for efficient policy training? A suitable generative model must capture the entire distribution of the training data to generate as many distinct motion trajectories as possible, while avoiding the generation of invalid trajectories outside the training dataset. 
The diversity of the generated data enables the policy to complete a given task for the entire set of goal states when training a goal-conditioned policy. On the other hand, adhering to the distribution of the training data ensures safety in generated trajectories which are running on a real robotic platform. 


In this paper, we (i) propose a learning framework that exploits RL and generative models to solve sequential decision making problems and considerably improves the data-efficiency of deep visuomotor policy training to control a robotic arm given raw image pixels as the input,
and (ii) provide a set of measures to evaluate the quality of the latent space of different generative models regulated by the RL policy search algorithms, and use them as a guideline for training the generative models such that the data-efficiency of the policy training can be further improved prior to actual training on a physical robot. 

Regarding (i), the proposed learning framework divides the deep visuomotor sequential decision-making problem into the following sub-problems that can be solved more efficiently:
(a) an unsupervised generative model training problem that approximates the distribution of motor actions, (b) a trust-region policy optimization problem that solves a contextual multi-armed bandit without temporal credit assignment issue which exists in typical sequential decision-making problems, and (c) a supervised learning problem in which we train the deep visuomotor policy in an end-to-end fashion.

Regarding (ii), we evaluate generative models based on (a) the quality and coverage of the samples they generate using the precision and recall metric \cite{kynkaanniemi2019improved}, and (b) the quality of their latent representations using two novel measures called \textit{disentangling precision and recall (DPR)} and \textit{latent local linearity (L3)}. Both these measures leverage the end states obtained after execution of the generated trajectories on a robotic platform. Disentanglement measures to which extent individual dimensions in the latent space control different aspects of the task, while local linearity measures the complexity of the generative process and system dynamics in the neighbourhood of each point in the latent space. Our hypothesis is that a generative model that is well disentangled, locally linear and able to generate realistic samples that closely follow the training data (i.e. has high precision and high recall) leads to a more sample efficient 
neural network policy training. We experimentally investigate this hypothesis on several generative models, namely $\beta$-VAEs \cite{higgins2017beta} and InfoGANs \cite{chen2016infogan}, by calculating Pearson's R as well as automatic relevance determination regression (ARD) to quantify the importance of (a) and (b) for a superior RL policy training performance. This evaluation provides a guideline for training latent-variable generative models in a way that enables data-efficient policy training.

In summary, the advantages of the proposed framework are: 
\begin{itemize}
    \item It improves data-efficiency of the policy training algorithm by at least an order of magnitude by incorporating prior knowledge in terms of a distribution over valid sequences of actions, therefore, reducing the search space. 
    \item It helps to acquire complex visuomotor policies given sparse terminal rewards provided at the end of successful episodes. The proposed formulation converts the sequential decision-making problem into a contextual multi-armed bandit. Therefore, it alleviates the temporal credit assignment problem that is inherent in  sequential decision-making tasks and enables efficient policy training with only terminal rewards.  
    \item It enables safe exploration in RL by sampling actions only from the approximated distribution. 
    This is in stark contrast to the typical RL algorithms in which random actions are taken during the exploration phase. 
    
    \item It provides a set of measures for evaluation of the generative model based on which it is possible to predict the performance of the RL policy training prior to the actual training.
\end{itemize}

This paper provides a comprehensive overview of our earlier work for RL policy training based on generative models \cite{ghadirzadeh2017deep, arndt2019meta, chen2019adversarial,hamalainen2019affordance, butepage2019imitating} and is organized as follows: in Section \ref{sec:related_work}, we provide an overview of the related work. We formally introduce the problem of policy training with generative models in Section \ref{sec:preliminaries}, and describe how the framework is trained in Section \ref{sec:em_policy_training}. In Section \ref{sec:generative_model_training} we first briefly overview VAEs and GANs, and then define all of the evaluation measures used to predict the final policy training performance.  
We present the experimental results in Section \ref{sec:experiments} and discuss the conclusion and future work in Section~\ref{sec:conclusion}.
Moreover, for the sake of completeness, we describe the end-to-end training of the perception and control modules in Appendix \ref{sec:perception} by giving a summary of the prior work \cite{levine2016end, chen2019adversarial}. Note that this work provides a complete overview of the proposed framework and focuses on the evaluation of the generative model. We refer the reader to \cite{ghadirzadeh2017deep} for investigation of the data-efficiency of the proposed approach in training complex visuomotor skills.

%% file: inputs/related_work.tex
\section{Related work}
\label{sec:related_work}

Our work addresses two types of problems: 
(a) visuomotor policy training based on unsupervised generative model training and trust-region policy optimization, and
(b) evaluation of generative models to forecast the efficiency of the final policy training task. 
We introduce the related work for each of the problems in the following sections. 

\textbf{Data-efficient end-to-end policy training:}
In recent years, end-to-end training of visuomotor policies using deep RL has gained in popularity in robotics research \cite{ghadirzadeh2017deep, levine2016end, finn2016deep, kalashnikov2018qt, quillen2018deep, singh2017gplac, devin2018deep, pinto2017asymmetric}. 
However, deep RL algorithms are typically data-hungry and learning a general policy, i.e., a policy that performs well also for previously unseen inputs, requires a farm of robots continuously collecting data for several days \cite{levine2018learning, finn2017deep, gu2017deep, dasari2019robonet}.
The limitation of large-scale data collection has hindered the applicability of RL solutions to many practical robotics tasks. Recent studies tried to improve
the data-efficiency by training the policy in simulation and transferring the acquired visuomotor skills to the real setup \cite{quillen2018deep, pinto2017asymmetric, abdolmaleki2020distributional, peng2018sim}, a paradigm known as sim-to-real transfer learning. 
Sim-to-real approaches are utilized for two tasks in deep policy training: (i) training the perception model via randomization of the texture and shape of visual objects in simulation and using the trained model directly in the real world setup (zero-shot transfer) \cite{hamalainen2019affordance, tobin2017domain}, and (ii) training the policy in simulation by randomizing the dynamics of the task and transferring the policy to the real setup by fine-tuning it with the real data (few-shot transfer learning) \cite{arndt2019meta, peng2018sim}. 
However, challenges in the design of the simulation environment can cause large differences between the real and the simulated environments which hinder an efficient knowledge transfer between these two domains. 
In such cases, transfer learning from other domains, e.g., human demonstrations \cite{butepage2019imitating, yu2018one} or simpler task setups \cite{chen2019adversarial, chen2018deep}, can help the agent to learn a policy more efficiently. 
In this work, we exploit human demonstrations to shape the robot motion trajectories by training generative models that reproduce the demonstrated trajectories. 
Following our earlier work \cite{chen2019adversarial}, we exploit adversarial domain adaptation techniques \cite{tzeng2017adversarial, tzeng2020adapting} to improve the generality of the acquired policy when it is trained in a simple task environment with a small amount of training data. 
In the rest of this section, we review related studies that improve the data-efficiency and generality of RL algorithms by utilizing trust-region terms, converting the RL problem into a supervised learning problem, and trajectory-centered approaches that shape motion trajectories prior to the policy training.

Improving the policy while avoiding abrupt changes in the policy distribution after each update is known as the trust-region approach in policy optimization. 
Trust-region policy optimization (TRPO) \cite{schulman2015trust} and proximal policy optimization (PPO) \cite{schulman2017proximal} are two variants of trust-region policy gradient methods that scale well to non-linear policies such as neural networks.
The key component of TRPO and PPO is a surrogate objective function with a trust-region term based on which the policy can be updated and monotonically improved. 
In TRPO, the changes in the distributions of the policies before and after each update are penalized by a KL divergence term. 
Therefore, the policy is forced to stay in a trust-region given by the action distribution of the current policy. 
Our EM formulation yields a similar trust-region term with the difference being that it penalizes the changes in the distribution of the deep policy and a so-called variational policy that will be introduced as a part of our proposed optimization algorithm. Since our formulation allows the use of any policy gradient solution, we use the same RL objective function as in TRPO. 

The EM algorithm has been used for policy training in a number of prior work \cite{neumann2011variational ,deisenroth2013survey,levine2013variational}. The key idea is to introduce variational policies to decompose the policy training into two downstream tasks that are trained iteratively until no further policy improvement can be observed \cite{ghadirzadeh2018sensorimotor}. 
In \cite{levine2016end} the authors introduced the guided policy search (GPS) algorithm which divides the visuomotor policy training task into a trajectory optimization and a supervised learning problem. 
GPS alternates between two steps: (i) optimizing a set of trajectories by exploiting a trust-region term to stay close to the action distribution given by the deep policy, and (ii) updating the deep policy to reproduce the motion trajectories. 
Our EM solution differs from the GPS framework and earlier approaches in that we optimize the trajectories by regulating a generative model that is trained prior to the policy training. 
Training generative models enables the learning framework to exploit human expert knowledge as well as to optimize the policy given only terminal rewards as explained earlier.

Trajectory-centric approaches, such as dynamic movement primitives (DMPs), have been popular because of the ease of integrating expert knowledge in the policy training process via physical demonstration \cite{peters2006policy, peters2008reinforcement, ijspeert2003learning, ijspeert2013dynamical, hazara2019transferring}.
However, such models are less expressive compared to deep neural networks and are particularly limited when it comes to end-to-end training of the perception and control elements of the model. Moreover, these approaches cannot be used to train reactive policies where the action is adjusted in every time-step based on the observed sensory input \cite{haarnoja2018composable}. 
On the other hand, deep generative models can model complex dependencies within the data by learning the underlying data distribution from which realistic samples can be obtained. 
Furthermore, they can be easily accommodated in larger neural networks without affecting the data integrity. Our framework based on generative models enables training both feedback (reactive) and feedforward policies by adjusting the policy network architecture. 

The use of generative models in robot learning has become popular in recent years \cite{ghadirzadeh2017deep, butepage2019imitating, hamalainen2019affordance, chen2019adversarial, arndt2019meta, lippi2020latent, gothoskar2020learning, igl2018deep, buesing2018learning, mishra2017prediction, ke2018modeling, hafner2018learning, rhinehart2018deep, krupnik2019multi} because of their  low-dimensional and regularized latent spaces. 
However, latent variable generative models are mainly studied to train a long-term state prediction model that is used in the context of trajectory optimization and model-based reinforcement learning  \cite{buesing2018learning, mishra2017prediction, ke2018modeling, hafner2018learning, rhinehart2018deep, krupnik2019multi}.
Regulating generative models based on reinforcement learning to produce sequences of actions according to the  visual state has first appeared in our prior work \cite{ghadirzadeh2017deep}. Since then we applied the framework in different robotic tasks, e.g., throwing balls \cite{ghadirzadeh2017deep}, shooting hockey-pucks \cite{arndt2019meta}, pouring into mugs \cite{chen2019adversarial, hamalainen2019affordance}, and in a variety of problem domains, e.g., sim-to-real transfer learning \cite{hamalainen2019affordance, arndt2019meta} and domain adaptation to acquire general policies \cite{chen2019adversarial}. 

\textbf{Evaluation of generative models:}
Although generative models have proved successful in many domains \cite{lippi2020latent, brock2018large, wang2018high, vae_anom, vae_text8672806} assessing their quality remains a challenging problem \cite{challenging_common}. It involves analysing the quality of both latent representations and generated samples. 
Regarding the latter, generated samples and their variation should resemble those obtained from the training data distribution. Early developed metrics such as IS \cite{IS_NIPS2016_6125}, FID \cite{FID_NIPS2017_7240} and KID \cite{binkowski2018demystifying} provided a promising start but were shown to be unable to separate between failure cases, such as mode collapse or unrealistic generated samples \cite{sajjadi2018assessing, kynkaanniemi2019improved}. Instead of using a one-dimensional score, \cite{sajjadi2018assessing} proposed to evaluate the learned distribution by comparing the samples from it with the ground truth training samples using the notion of precision and recall. Intuitively, precision measures the similarity between the generated and real samples, while recall determines the fraction of the true distribution that is covered by the distribution learned by the model. The measure was further improved both theoretically and practically by \cite{revisiting_pr}, while \cite{kynkaanniemi2019improved} provides an explicit non-parametric variant of the original probabilistic approach. We complement our measures for assessing disentanglement and local linearity of the latent representations with the precision and recall measure provided by \cite{kynkaanniemi2019improved}.

Regarding the assessment of the quality of the latent representation,
a widely adopted approach 
is the measure of disentanglement \cite{higgins2018towards, repr_learning_survey, tschannen2018recent}. A representation is said to be disentangled if each latent component encodes exactly one ground truth generative factor present in the data \cite{kim2018disentangling}. Existing frameworks for both learning and evaluating disentangled representations \cite{higgins2017beta, kim2018disentangling, eastwood2018framework, chen2018isolating,kumar2017variational} rely on the assumption that the ground truth factors of variation are known a priori and are independent. The core idea is to measure how changes in the generative factors affect the latent representations and vice versa. In cases when an encoder network is available, this is typically achieved with a classifier that was trained to predict which generative factor was held constant given a latent representation \cite{higgins2017beta, kim2018disentangling, eastwood2018framework, kumar2017variational, chen2018isolating}. In generative models without an encoder network, such as GANs, disentanglement is measured by visually inspecting the latent traversals provided that the input data are images \cite{chen2016infogan, jeon2019ibgan, lee2020high, liu2019oogan}. However, these measures are difficult to apply when generative factors of variation are unknown or when manual visual inspection is not possible,
both of which is the case with sequences of motor commands for controlling a robotic arm. We therefore define a measure of disentanglement that does not rely on any of these requirements and instead
leverages the end states of the downstream robotics task corresponding to a given set of latent action representations. In contrast to existing measures it measures how changes in the latent space affect the obtained end states in a fully unsupervised way. Moreover, since the generative model in our case is combined with the system dynamics, we complement the evaluation of the latent representations with a measure of local linearity which quantifies the complexity of the system dynamics in a neighbourhood of a given latent sample.


%% file: inputs/background.tex
\section{Preliminaries}
\label{sec:preliminaries}

\begin{figure}[ht]
\centering
\includegraphics[width=0.7\linewidth]{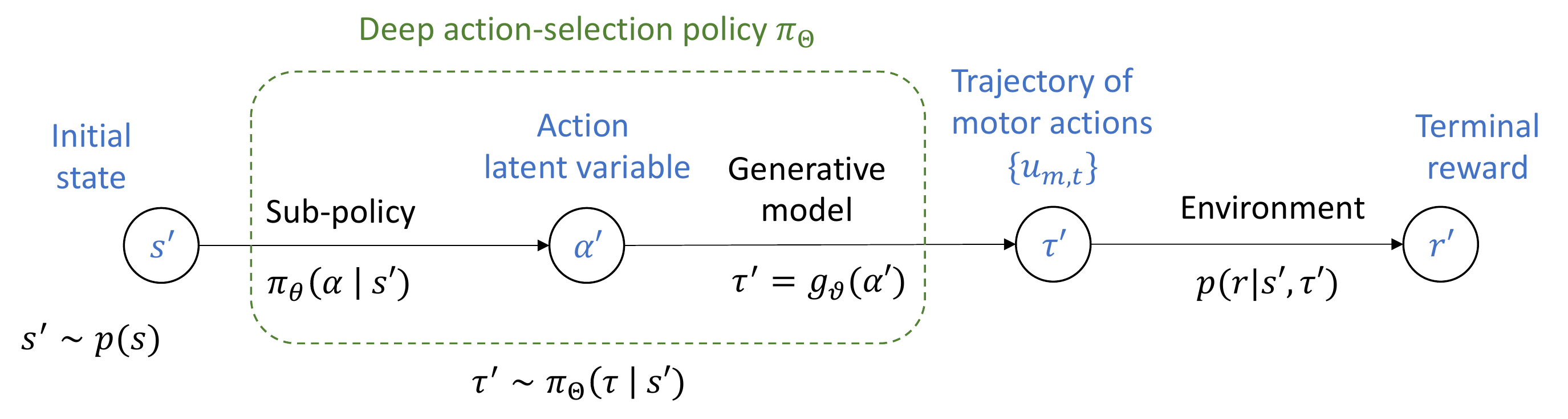} 
\caption{The architecture of the deep action-selection policy $\pi_\Theta$ based on latent-variable generative models. The policy consists of two models, the sub-policy $\pi_\theta(\alpha|s')$ that assigns a distribution over the action latent variable $\alpha$ conditioned on a given state $s'$, and a generative model $\tau' = g_\vartheta(\alpha')$ that maps a latent action sample $\alpha' \sim \pi_\theta(\alpha|s')$  into a trajectory of motor actions $\tau'$. Given the initial state $s'$ and the action trajectory $\tau'$ that is executed on the robot, the environment returns a terminal reward $r'$ according to the reward probability $r' \sim p(r|s',\tau')$.}\label{fig:deep_policy_network}
\end{figure}

We consider a finite-horizon Markov decision process defined by a tuple $(\mathbf{S}, \mathbf{U}, P, p(s), p(r|s,\tau))$ where $\mathbf{S} = \mathbb{R}^{N_s}$ is a set of $N_s$-dimensional states, $\mathbf{U}$ is a set of motor actions $u_{m}$ for the robot motor $m$, 
$P: \mathbf{S} \times \mathbf{U} \times \mathbf{S} \rightarrow \mathbb{R}$ is the state transition probability, and $p(s)$ is the initial state distribution. The probability $p(r|s',\tau')$ of the reward $r$, conditioned on a state $s'$ and a fixed-length sequence $\tau' = (u_{t, m})$ of motor $m$ actions $u_{t, m}$ at time step $t$, is unknown to the learning agent.
We wish to find a policy $\pi_\Theta(\tau|s')$ 
based on which a sequence of motor actions $\tau'$ can be sampled given a state of the environment $s'$.
The state contains information about the current configuration of the environment as well as the goal to reach in case a goal-conditioned policy has to be obtained. 
Let $r^*$ be the minimum reward required to successfully complete the task.
The policy $\pi_\Theta$ is represented by a neural network with parameters $\Theta$ that are trained to maximize the probability of receiving high rewards $p(r \ge r^*|s', \tau')$ 
\begin{equation}
\begin{split}
    \Theta^* & = \argmax_{\Theta} \iint p(s)\, \pi_{\Theta}(\tau | s)\, p(r|s,\tau)\, ds \, d\tau \\
     & = \argmax_{\Theta} \mathbb{E}_{s' \sim p(s), \tau' \sim \pi_{\Theta}(\tau|s')} [\, p(r | s', \tau')\,],
    \label{eq:margin_tau}
\end{split}
\end{equation}
where we omitted the reward threshold $r^*$ for simplicity.
Our approach is based on training a generative model $g_\vartheta$, parametrized by $\vartheta$, that maps a low-dimensional latent action sample $\alpha' \in \mathbb{R}^{N_\alpha}$ into a motion trajectory $\tau' \in \mathbb{R}^{T\times M}$, where $N_\alpha \ll T\times M$. 
In other words, we assume that the search space is limited to the trajectories spanned by the generative model. 
In this case, the feed-forward policy search problem splits into two sub-problems: (i) finding the mapping $g_\vartheta(\alpha')$ and (ii) finding the sub-policy $\pi_\theta (\alpha | s')$, where $\Theta = [\theta, \vartheta]$. 
Instead of marginalizing over trajectories as in~\eqref{eq:margin_tau} we marginalize over the latent variable by exploiting the generative model
\begin{equation}
\theta^* = \argmax_{\theta} \mathbb{E}_{s' \sim p(s), \alpha' \sim \pi_{\theta}(\alpha|s')} [\, p(r|s', g_\vartheta(\alpha')\,)\,].
\label{eq:margin_alpha}
\end{equation}
An overview of our approach is shown in Figure~\ref{fig:deep_policy_network}. Once the models are trained the output of the policy $\pi_\Theta$ is found by first sampling from the sub-policy $\alpha' \sim \pi_\theta(\alpha | s')$ given a state $s'$ and then using the mapping $g_\vartheta$ to get the sequence of motor actions $\tau' = g_\vartheta(\alpha')$. The state $s'$ and the generated trajectory $\tau'$ are then given to the environment which outputs a reward $r'$.
In the rest of the text we refer to the sub-policy as the \textit{policy} and omit the parameters $\vartheta$ from the notation $g_\vartheta$ when they are not needed.
By abuse of notation, we will drop the explicit distinction between a random variable, e.g., $\alpha$, and a concrete instance of it, $\alpha'$, in the rest of the paper and only write $\alpha$ when no confusions can arise. 
In the following section, we introduce the expectation-maximization (EM) algorithm for training an action-selection policy using a generative model based on which different motor trajectories suitable for solving a given task can be generated. 

%% file: inputs/policy_training.tex
\section{Expectation-Maximization Policy Training}
\label{sec:em_policy_training}
The EM algorithm is a well-suited approach to find the maximum likelihood solution to the intractable marginalization over the latent variable introduced in~\eqref{eq:margin_tau} and~\eqref{eq:margin_alpha}. 
We use the EM algorithm to find an optimal policy $\pi_{\theta^*}(\alpha|s)$ by first introducing a variational policy $q(\alpha|s)$ which is a simpler auxiliary distribution used to improve the training of $\pi_{\theta}$. 
As the goal is to find an action trajectory $\tau$ that maximizes the reward probability $p(r|s,\tau)$, we start by expressing its logarithm as $\log p(r | s) = \int q(\alpha|s) \log p(r|s) d\alpha$, where we used the identity $\int q(\alpha|s) d\alpha = 1$ and omitted the conditioning on $\tau$ in the reward probability for simplicity.
Following the EM derivation introduced in \cite{neumann2011variational} and using the identity $p(r|s) = p(r,\alpha|s)/p(\alpha |r,s)$, the expression can be further decomposed into
\begin{align}
\log p(r|s) & = \underbrace{\int q(\alpha|s) \log \frac{p(r, \alpha|s)}{q(\alpha|s)} d\alpha}_{\text{I}}  + \underbrace{\int q(\alpha|s) \log \frac{q(\alpha|s)}{p (\alpha | r, s)}\,d\alpha}_{\text{II}} \label{eq:marginal_decomposed} 
\end{align}

The second term (II) is the Kullback-Leibler (KL) divergence $D_{KL}( q(\alpha|s) \,||\, p(\alpha |
r,s) )$ between distributions $q(\alpha|s)$ and $p(\alpha |
 r,s)$, which is a non-negative quantity.
Therefore, the first term (I) provides a lower-bound for $\log p(r|s)$. To maximize the latter we use the EM algorithm which is an iterative procedure consisting of two steps known as the expectation (E-) and the maximization (M-) steps, introduced in the following sections. 

\subsection{Expectation step } 
\label{sec:EM}
The E-step yields a trust-region policy optimization objective which solves a contextual multi-armed bandit without temporal complexities. The objective of the E-step is to minimize the KL divergence term (II) in~\eqref{eq:marginal_decomposed} by optimizing $q(\alpha | s)$, which in turn indirectly maximizes the lower bound (I).
Since $\log p(r|s)$ does not depend on $q(\alpha|s)$ the sum of the KL divergence term (II) and the lower bound term (I) is a constant value for different $q$. Therefore, reducing (II) by optimizing $q$ increases the lower bound (I). Assuming that $q$ is parametrized by $\phi$, the E-step objective function is given by
\begin{align}
    \phi^* &=\argmin_{\phi} D_{KL}(\,q_\phi(\alpha|s)\,||\,p(\alpha|r,s)\,) \nonumber \\
     &= \argmax_{\phi} \mathbb{E}_{\alpha' \sim q_\phi(\alpha|s)}[\log p(r|s, \alpha')] -  D_{KL}(\,q_\phi(\alpha|s)\,||\,\pi_\theta(\alpha|s)\,),
    \label{eq:E_loss}
\end{align}
where we used the Bayes rule $p(\alpha|r,s) =  p(r|\alpha,s) p(\alpha|s)/p(r|s)$ and substituted $p(\alpha|s)$ by $\pi_\theta (\alpha|s)$. 
In typical RL applications, we maximize the reward value given by a stochastic reward function $r(s,\tau)$. 
In this case, $\mathbb{E}_{q_\phi(\alpha|s)}[\log p(r| s, \alpha)]$ can be maximized indirectly by maximizing the expected reward value $\mathbb{E}_{q_{\phi}(\alpha|s)}[r(s, \alpha)]$ on which we can apply the policy gradient theorem. Note that by $r(s, \alpha)$ we refer to $r(s, g_\vartheta(\alpha))$.
Moreover, $D_{KL}(\,q_\phi(\alpha|s)\,||\,\pi_\theta(\alpha|s)\,)$ acts as a trust region term forcing $q_\phi$ not to deviate too much from the policy distribution $\pi_\theta$. 
Therefore, we can apply policy search algorithms with trust region terms to optimize the objective given in~\eqref{eq:E_loss}. Following the derivations introduced in \cite{schulman2015trust}, we adopt TRPO objective for the E-step optimization
\begin{equation}
    \phi^* = \argmax_{\phi} \mathbb{E}_{s' \sim p(s), \alpha' \sim \pi_\theta(\alpha|s')} \left[\frac{q_\phi(\alpha'|s')}{\pi_\theta(\alpha'|s')}\,A(s', \alpha') - D_{KL}(q_\phi(\alpha|s') \, || \, \pi_\theta(\alpha | s'))\right ], 
    \label{eq:trpo}
\end{equation}
where $A(s', \alpha') = r(s', \alpha') - V_\pi(s')$ is the advantage function, $V_\pi(s') = \mathbb{E}_{ \alpha' \sim \pi_\theta(\alpha|s')} [r(s', \alpha')]$ is the value function and $\phi^*$ denotes the optimal solution for the given iteration. Note that the action latent variable $\alpha$ is always sampled from the policy $\pi_\theta(\alpha|s)$ and not from the variational policy $q_\phi(\alpha|s)$.


\subsection{Maximization step }
The M-step yields a supervised learning objective using which we train the deep policy in an end-to-end fashion. It directly maximizes the lower bound (I) in~\eqref{eq:marginal_decomposed} by optimizing the policy parameters $\theta$ while holding the variational policy $q_\phi$ constant. Following \cite{deisenroth2013survey} and noting that the dynamics of the system $p(r|\alpha, s)$ are not affected by the choice of the policy parameters $\theta$, we maximize (I) by minimizing the following KL divergence
\begin{equation}
\theta^* = \argmin_{\theta} D_{KL}(\,q_\phi(\alpha|s)\, || \,\pi_{\theta}(\alpha|s)\,).
\label{eq:M_loss}
\end{equation}

In other words, the M-step updates the policy $\pi_{\theta}$ to match the distribution of the variational policy $q_\phi$ which was updated in the E-step. Similarly as in the E-step, $\theta^*$ denotes the optimal solution for the given iteration. The M-step can be combined with end-to-end training of the perception and control modules. We refer the reader to the Appendix \ref{sec:perception} for the details.

A summary of the EM policy training is given in Algorithm \ref{alg:training}. In each iteration, a set of states $\{s_i\}$ is sampled from the initial state distribution $p(s)$. For each state $s_i$, a latent action sample $\alpha_i$ is sampled from the distribution given by the policy $\pi_\theta(\alpha|s_i)$. A generative model $g$ is then used to map every latent action variable $\alpha_i$ into a full motor trajectory $\tau_i$ which is then deployed on the robot to get the corresponding reward value $r_i$. 
In the inner loop, the variational policy $q_\phi$ and the main policy $\pi_\theta$ are updated iteratively based on gradient descent on batches of data using the objective function for the E- and M-steps of the policy optimization method.
\input{inputs/algorithm}


%% file: inputs/algorithm.tex
\begin{algorithm}[h]
\SetKwInOut{Input}{Input}
\SetKwInOut{Output}{Output}
\caption{Pseudocode of the EM policy training based on generative models.}
\Input{generative model $g_\vartheta$, initial policy $\pi_\theta$, initial value function $V_\pi$, batch size $N$}
\Output{trained $\pi_\theta$}
\While{training $\pi_\theta$}{
\For{$i = 1, \dots, N$}{
sample states $s_i \sim p(s)$ \\
sample actions $\alpha_i \sim \pi_\theta(.|s_i)$ \\
generate motor actions $\tau_i \leftarrow g_\vartheta(\alpha_i)$  \\
obtain the rewards $r_i\leftarrow r(s_i, \tau_i)$ \\
}
\Repeat{training done}{
\textbf{E-step:} \\
\quad update the variational policy $q_\phi$ according to~\eqref{eq:trpo} given $\{s_i, \alpha_i, r_i\}_{i=1}^N$ \\
\textbf{M-step:} \\
\quad update the policy $\pi_\theta$ according to~\eqref{eq:M_loss} given $q_\phi$ and $\{s_i, \alpha_i\}_{i=1}^N$\\
}
update the value function $V_\pi$ given $\{s_i,r_i\}_{i=1}^N$\\
}
\label{alg:training}
\end{algorithm}

%% file: inputs/generative_model_training.tex
\section{Generative Model Training}
\label{sec:generative_model_training}
So far we discussed how to train an action-selection policy based on the EM algorithm to regulate the action latent variable which is the input to a generative model. 
In this section, we review two prominent approaches to train a generative model, Variational Autoencoder (VAE) and Generative Adversarial Network (GAN), which we use to generate sequences of actions required to solve the sequential decision-making problem. 
We then introduce a set of measures used to predict which properties of a generative model will influence the performance of the policy training. 

\subsection{Training generative models}
We aim to model the distribution $p(\tau)$ of the motor actions  that are suitable to complete a given task. To this end, we introduce a low-dimensional random variable $\alpha$ with a probability density function $p(\alpha)$ representing the latent actions which are mapped into unique trajectories $\tau$ by a generative model $g$. The model $g$ is trained to maximize the likelihood $\mathbb{E}_{\tau \sim \mathcal{D}, \alpha' \sim p(\alpha)}[p_\vartheta(\tau|\alpha')]$ of the training trajectories $\tau \in \mathcal{D}$ under the entire latent variable space.

\subsubsection{Variational autoencoders}
A VAE \cite{kingma2014auto, rezende2014stochasticvae2} consists of encoder and decoder neural networks representing the parameters of the approximate posterior distribution $q_\varphi(\alpha | \tau)$ and the likelihood function $p_\vartheta(\tau|\alpha)$, respectively. The encoder and decoder neural networks, parametrized by $\varphi$ and $\vartheta$, respectively, are jointly trained to optimize the variational lower bound
\begin{equation}
     \max_{\varphi, \vartheta} \mathbb{E}_{\alpha' \sim q_\varphi(\alpha|\tau)}[\log p_\vartheta(\tau|\alpha')] - \beta D_{KL}(q_\varphi (\alpha | \tau) || p(\alpha)),
\label{eq:vae}
\end{equation}
where the prior $p(\alpha)$ is the standard normal distribution and the parameter $\beta$ \cite{higgins2017beta} a variable controlling the trade-off between the reconstruction fidelity and the structure of the latent space regulated by the KL divergence. A $\beta > 1$ encourages the model to learn more disentangled latent representations \cite{higgins2017beta}.

\subsubsection{Generative adversarial networks}
A GAN model \cite{goodfellow2014generative} consists of a generator and discriminator neural networks that are trained by playing a min-max game. The generative model $g_\vartheta$, parametrized by $\vartheta$, transforms a latent sample $\alpha'$ sampled from the prior noise distribution $p(\alpha)$ into a trajectory $\tau = g_\vartheta(\alpha')$. The model needs to produce realistic samples resembling those obtained from the training data distribution $p(\tau)$. It is trained by playing an adversarial game against the discriminator network $D_\varphi$, parametrized by $\varphi$, which needs to distinguish a generated sample from a real one. The competition between the two networks is expressed as the following min-max objective
\begin{align}
    \min_\vartheta \max_\varphi \mathbb{E}_{\tau' \sim p(\tau)} [\log D_\varphi(\tau')] + \mathbb{E}_{\alpha' \sim p(\alpha)} [\log(1 - D_\varphi(g_\vartheta(\alpha'))]. \label{eq:gan_original}
\end{align}
However, the original GAN formulation~\eqref{eq:gan_original} does not impose any restrictions on the latent variable $\alpha$ and therefore the generator $g_\vartheta$ can use $\alpha$ in an arbitrary way. To learn disentangled latent representations 
we instead use InfoGAN \cite{chen2016infogan} which is a version of GAN with an information-theoretic regularization added to the original objective. The regularization is based on the idea to maximise the mutual information $I(\alpha', g_\vartheta(\alpha'))$ between the latent code $\alpha'$ and the corresponding generated sample $g_\vartheta(\alpha')$. 
An InfoGAN model is trained using the following information minmax objective \cite{chen2016infogan}
\begin{equation}
    \min_{\vartheta, \psi} \max_\varphi \mathbb{E}_{\tau' \sim p(\tau)} [\log D_\varphi(\tau')] + \mathbb{E}_{\alpha' \sim  p(\alpha)} [\log(1 - D_\varphi(g_\vartheta(\alpha'))] - \lambda \mathbb{E}_{\alpha' \sim p(\alpha), \tau' \sim g_\vartheta(\alpha')}[\log Q_\psi(\alpha' | \tau')],
\label{eq:gan}
\end{equation}
where $Q_\psi(\alpha | \tau')$ is an approximation of the true unknown posterior distribution $p(\alpha | \tau')$ and $\lambda$ a hyperparameter. In practice, $Q_\psi$ is a neural network that models the parameters of a Gaussian distribution and shares all the convolutional layers with the discriminator network $D_\varphi$ except for the last few output layers.



\subsection{Evaluation of the generative models}
\label{sec:eval_generative_model}
We review the characteristics of generative models that can potentially improve the policy training by measuring precision and recall, disentanglement and local linearity. Our goal is to be able to judge the quality of the policy training by evaluating the generative models prior to the RL training. We relate the measures to the performance of the policy training in Section~\ref{sec:exp:generative_model}.

\subsubsection{Disentangling precision and recall}
In this section, we define our measure, called \textit{disentangling precision and recall,} for evaluating the disentanglement of latent action representations. 
A disentangled representation of the motor data obtained from the latent space of a generative model can be defined as the one in which every end state of the system is controllable by one latent dimension determined by the vector basis of the latent space. 
For example, consider the picking task where the goal is to pick an object on a table top (Section~\ref{sec:exp:setup}). We say that a latent representation given by a generative model is well-disentangled if there exists a basis of the latent space with respect to which each dimension controls one axis of the position of the end-effector.
Our hypothesis is that the more disentangled the representation is, the more efficient is the policy training. 
While disentangled representations have been successfully applied to a variety of downstream machine learning tasks \cite{disentanglement_video, creager2019flexibly, lee2018diverse}, their usefulness has been questioned by \cite{challenging_common} and \cite{van2019disentangled} who have observed disagreements among the existing disentanglement metrics. 
We experimentally evaluate the effect of disentangled representations on the performance of the policy training in Section~\ref{sec:exp:eval_gen_models}.



Our disentanglement measure is based on statistical testing performed on the end state space of the system.
Let $\boldsymbol{S}_r$ be the set of end states obtained by executing the training motor trajectories on a robotic platform. If representations given by $g$ are well disentangled, then setting one latent dimension to a fixed value should result in limited variation in the corresponding generated end states $\boldsymbol{S}_g$. For example, if the $1$st latent dimension controls the $x$-axis position of the end-effector in the picking task then setting it to a fixed value should limit the set of possible $x$ positions. 
In other words, we wish to quantify how dissimilar the set of end states $\boldsymbol{S}_g$, obtained by holding one latent dimension constant, is from the set $\boldsymbol{S}_r$. To compute such dissimilarity we use maximum mean discrepancy (MMD) \cite{JMLR:v13:gretton12a} which is a statistical test for determining if two sets of samples were produced by the same distribution. Using kernels, MMD maps both sets into a feature space called reproducing kernel Hilbert space, and computes the distance between mean values of the samples in each group. In our implementations, we compute the unbiased estimator of the squared MMD (Lemma 6 in \cite{JMLR:v13:gretton12a}) given by 
\begin{align*}
    \text{MMD}^2(\boldsymbol{S}_r, \boldsymbol{S}_g) = \frac{1}{m (m -1)} \sum_{i \neq j}^m k(s_r^i, s_r^j) + \frac{1}{n (n -1)} \sum_{i \neq j}^n k(s_g^i, s_g^j) - \frac{2}{mn} \sum_{i = 1}^m \sum_{j = 1}^n k(s_r^i, s_g^j),
\end{align*}
where, $\boldsymbol{S}_r = \{s_r^1, \dots, s_r^m\}$, $\boldsymbol{S}_g = \{s_g^1, \dots, s_g^n\}$ are the two sets of samples and $k(x, y) = \exp(- \gamma ||x - y||^2)$ is the exponential kernel with hyperparameter $\gamma$ determining the smoothness. Due to the nature of the exponential kernel, the higher the MMD score, the lower the similarity between the two sets.
 
\begin{figure}[h]
\centering
\includegraphics[width=0.99\linewidth]{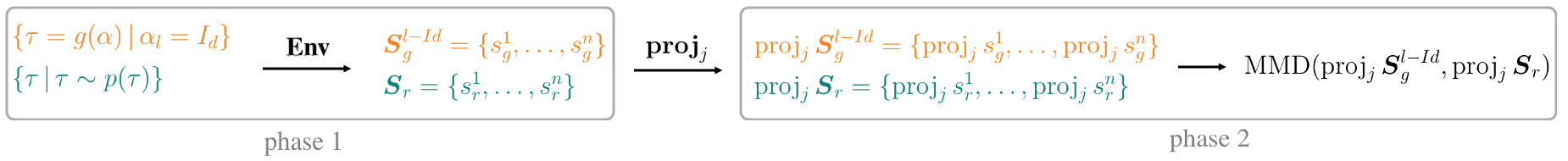}
\caption{Visualisation of phase 1 and 2 of our disentangling precision and recall metric $\Dis$. }\label{fig:disentanglement_graphics}
\end{figure}

Our measure can be described in three phases. We provide an intuitive summary of each phase but refer the reader to Appendix \ref{app:dis_score} for a rigorous description of the algorithm. In phase $1$ (Figure \ref{fig:disentanglement_graphics} left) we generate the two sets of end states on which we run the statistical tests. For a fixed latent dimension $l \in \{1, \dots, N_\alpha\}$ we perform a series of $D \in \mathbb{N}$ \textit{latent interventions} where we set the $l$th component of a latent code $\alpha_l$ to a fixed value $I_d$, $\alpha_l = I_d$ for $d = 1, \dots, D$. Each intervention $\alpha_l = I_d$ is performed on a set of $n$ samples sampled from the prior distribution $p(\alpha)$. We denote by $\boldsymbol{S}_g^{l-I_d}$ the set of $n$ end states obtained by executing the trajectories generated from the latent samples on which we performed the intervention $\alpha_l = I_d$. For example, $D = 5$ latent interventions on the $1$-st latent dimension yield the sets $\boldsymbol{S}_g^{1-I_1}, \dots, \boldsymbol{S}_g^{1-I_5}$. Moreover, we denote by $\boldsymbol{S}_r$ the set of $n$ randomly subsampled end states that correspond to the training motor data.  

In phase $2$ (Figure \ref{fig:disentanglement_graphics} right), we perform the $\MMD$ tests on each pair of sets $\boldsymbol{S}_g^{l-I_d}$ and $\boldsymbol{S}_r$ obtained in phase $1$. In particular, we wish to determine if an intervention on a given dimension $l$ induced a change in any of the components $j = 1, \dots, N_s$ of the end state space. If such a change exists, we consider the latent dimension $l$ to be well disentangled. Moreover, if we can find a set of latent dimensions that induce changes on different components of the end states, we consider the generative model $g$ to be well disentangled. Therefore, for a fixed latent dimension $l$ and a fixed intervention $\alpha_l = I_d$, the objective is to find the component $j$ of the end state space for which $\boldsymbol{S}_g^{l-I_d}$ and $\boldsymbol{S}_r$ are most dissimilar. This translates to finding the component $j$
yielding the largest 
value $\MMD(\proj_j \boldsymbol{S}_g^{l-I_d}, \proj_j \boldsymbol{S}_r)$  where $\proj_j \boldsymbol{S} = \{ \proj_j s \,|\,s \in \boldsymbol{S} \}$ denotes the set of the $j$th components of the states from the set $\boldsymbol{S}$. Note that if the dimension $l$ is entangled such component $j$ does not exists (see Appendix \ref{app:dis_score} for details). 

In phase $3$ we aggregate the values of the performed $\MMD$ tests and define the final disentanglement score for the generative model $g$. In phase $2$ we linked each latent dimension $l$ with zero or one end state component $j$, and computed the corresponding value of the $\MMD$ test. Here, we first select $\min(N_s, N_\alpha)$ such pairs of latent space and end state space dimensions that yield the largest $\MMD$ values. In other words, we select the latent dimensions for which the latent interventions resulted in the largest changes in the end state space. We define $\Dip(g)$ as the sum of the selected $\MMD$ values, and $\Dir(g)$ as the number of unique end state space components present in the selected pairs, normalised by the total number of components $N_s$. Finally, we define the \textit{Disentangling Precision and Recall} $\Dis$ as a pair $$\Dis(g) = (\Dip(g), \Dir(g)).$$
Intuitively, $\Dip(g)$ quantifies the sum effect of the latent interventions with the most impact on the end states, and can therefore be thought of as \textit{disentangling precision}. A high $\Dip$ value indicates that the intervened latent samples resulted in end states with significantly limited variation, which in turn means that the latent disentanglement has a high precision. On the other hand, $\Dir(g)$ measures how many different aspects of the end states are captured in the latent space, and can be thus thought of as \textit{disentangling recall}. A high $\Dir$ value indicates that more end state components are captured in the latent space, such that the latent disentanglement has a high recall. The defined score is a novel fully unsupervised approximate measure of disentanglement for generative models combined with the RL policy training. Its absolute values can however vary depending on the kernel parameter $\gamma$ determining its smoothness. Moreover, this measure is not to be confused with the precision and recall from Section \ref{sec:prec_and_recall} where the aim is to evaluate the quality of the generated samples as opposed to the quality of the latent representations.

\subsubsection{Latent Local linearity}
The linearity of system dynamics plays a vital role in control theory but has not been studied in the context of generative model training. 
The system dynamics govern the evolution of the states as the result of applying a sequence of motor actions to the robot. Our hypothesis is that a generative model integrated with the environment performs better in the policy training if it satisfies the local linearity property defined below.

Let the mapping $\text{Exe}: \mathbb{R}^{T \times M} \rightarrow \mathbb{R}^{N_s}$ correspond to the execution of motion actions on a robot and let $s = \text{Exe}(\tau) \in \mathbb{R}^{N_s}$ denote the end state obtained by executing actions $\tau \in \mathbb{R}^{T \times M}$. 
Let $N_\varepsilon(\alpha) = \{\alpha': ||\alpha' - \alpha||_2 < \varepsilon\}$ be the Euclidean $\varepsilon$-neighbourhood of a latent action $\alpha$.  Then the composition of the maps $\text{Exe} \circ g: \mathbb{R}^{N_\alpha} \rightarrow \mathbb{R}^{N_s}$ mapping from the action latent space to the end state of the system is considered \textit{locally linear} in the neighbourhood of $\alpha$ if there exists an affine transformation 
\begin{align}
f_\alpha:  N_\varepsilon(\alpha) \subset \mathbb{R}^{N_\alpha} \label{eq:aff_trans} &\longrightarrow \mathbb{R}^{N_s} \\ \alpha' &\longmapsto A\alpha' + b \nonumber
\end{align}
such that $\text{Exe}(g(\alpha')) = f_\alpha(\alpha')$ for every $\alpha' \in N_\varepsilon(\alpha)$. We define the \textit{latent local linearity (L3)} of a generative model $g$ to be the mean square error (MSE) of $f_{\alpha_i}$ obtained on $N_\varepsilon(\alpha_i)$ calculated on a subset of latent actions $\{\alpha_i\}$.  

\subsubsection{Precision and recall} \label{sec:prec_and_recall}
Precision and recall for distributions is a measure, first introduced by \cite{sajjadi2018assessing} and further improved by \cite{kynkaanniemi2019improved}, for evaluating the quality of a distribution learned by a generative model $g$. It is based on the comparison of samples obtained from $g$ with the samples from the ground truth reference distribution. In our case, the reference distribution is the one of the training motor trajectories. Intuitively, \textit{precision} measures the quality of the generated sequences of motor actions by quantifying how similar they are to the training trajectories. It determines the fraction of the generated samples that are realistic. On the other hand, \textit{recall} evaluates how well the learned distribution covers the reference distribution and it determines the fraction of the training trajectories that can be generated by the generative model. In the context of the policy training, we would like the output of $\pi_\Theta$ to be as similar as possible to the demonstrated motor trajectories. It is also important that $\pi_\Theta$ covers the entire state space as it must be able to reach different goal states from different task configurations. Therefore, the generative model needs to have both high precision and high recall scores. 

The improved measure introduced by \cite{kynkaanniemi2019improved} is based on an approximation of
manifolds of both training and generated data. In particular, given a set $\boldsymbol{T} \in \{\boldsymbol{T_r}, \boldsymbol{T_g}\}$ of either real training trajectories $\boldsymbol{T_r}$ or generated trajectories $\boldsymbol{T_g}$, the corresponding manifold is estimated by forming hyperspheres around each trajectory $\tau \in \boldsymbol{T}$ with radius equal to its $k$th nearest neighbour $\NN_k(\tau, \boldsymbol{T})$. To determine whether or not a given novel trajectory $\tau'$ lies within the volume of the approximated manifold we define a binary function
\[
 f(\tau', \boldsymbol{T}) =
  \begin{cases}
  1 & \text{if $||\tau' - \tau||_2 \le ||\tau - \NN_k(\tau, \boldsymbol{T})||_2$ for at least one $\tau \in \boldsymbol{T}$} \\
  0 & \text{otherwise.} 
  \end{cases}
\]
By counting the number of generated trajectories $\tau_g \in \boldsymbol{T}_g$ that lie on the manifold of the real data $\boldsymbol{T}_r$ we obtain the \textit{precision}, and similarly the \textit{recall} by counting the number of real trajectories $\tau_r \in \boldsymbol{T}_r$ that lie on the manifold of the generated data $\boldsymbol{T}_g$
\begin{align*}
    \text{precision}(\boldsymbol{T}_r, \boldsymbol{T}_g) = \frac{1}{|\boldsymbol{T}_g|} \sum_{\tau_g \in \boldsymbol{T}_g} f(\tau_g, \boldsymbol{T}_r) \quad \text{and} \quad
    \text{recall}(\boldsymbol{T}_r, \boldsymbol{T}_g) = \frac{1}{|\boldsymbol{T}_r|} \sum_{\tau_r \in \boldsymbol{T}_r} f(\tau_r, \boldsymbol{T}_g).
\end{align*}
In our experiments, we use the original implementation provided by \cite{kynkaanniemi2019improved} directly on the trajectories as opposed to their representations as suggested in the paper.

%% file: inputs/experiments.tex
\section{Experiments}
\label{sec:experiments}

In this section, we experimentally determine the characteristics of generative models that contribute to a more data-efficient policy training for a picking task performed on a real robotic platform. We first trained several $\beta$-VAE and InfoGAN models with various hyperparameters, and evaluated them by measuring disentanglement, local linearity as well as precision and recall introduced in Section~\ref{sec:eval_generative_model}. Using these models and the proposed EM algorithm presented in Section~\ref{sec:em_policy_training}, we then trained several RL policies and investigated the relation between the properties of the generative models and the performance of the policy.

Note that the experimental section of this work focuses solely on the evaluation of the generative models. Readers interested in the investigation of the data-efficiency of the proposed approach in training complex visuomotor skills are referred to \cite{ghadirzadeh2017deep}. Moreover, we emphasize that it is not meaningful to directly compare our approach to neither PPO nor GPS. Training a policy using PPO requires vast amounts of data, while GPS requires a reward at every time step instead of the terminal reward as in our case.

\subsection{Experimental setup}
\label{sec:exp:setup}
We applied our framework to a picking task in which a 7 degree-of-freedom robotic arm (ABB YuMi) must move its end-effector to different positions and orientations on a tabletop to pick a randomly placed object on a table. In this case, raw image pixel value are given as the input to the visuomotor policy. This task is a suitable benchmark to answer our research questions. 
First of all, the task requires feed-forward control of the arm over 79 time-steps to precisely reach a target position without any position feedback during the execution.
Therefore, precision is an important factor for this problem setup. 
Secondly, reaching a wide range of positions and orientations on the tabletop requires the generative model $g$ to generate all possible combinations of motor commands that bring the end-effector to every possible target position and orientation. This means that $g$ needs to have a high recall. Thirdly, it is straightforward to evaluate the disentanglement of the latent representations  as well as the local linearity of the dynamical system that is formed by $g$ and the robot kinematic model. 
Finally, this is a suitable task for end-to-end training, especially by exploiting the adversarial domain adaptation technique in \cite{chen2019adversarial} to obtain generality for the policy training task. 
Note that the applicability of our framework (up to minor differences) to a wide range of robotic problems has already been addressed by our prior work. In particular, we successfully evaluated it in several robotic task domains, e.g., ball throwing to visual targets \cite{ghadirzadeh2017deep}, shooting hockey pucks using a hockey stick \cite{arndt2019meta}, pouring into different mugs \cite{hamalainen2019affordance}, picking objects \cite{chen2019adversarial}, and imitating human greeting gestures \cite{butepage2019imitating}. 

We constructed a dataset containing sequences of motor actions using MoveIt planners \cite{coleman2014reducing}.  
We collected $15750$ joint velocity trajectories to move the end-effector of the robot from a home position to different positions and orientations on a tabletop. 
The trajectories were sampled at $10$Hz and trimmed to 79 time-steps ($7.8$ seconds duration) by adding zeros at the end of the joint velocities that are shorter than 79 time-steps. 
The target positions as well as orientations consisting of Euler angles form a $N_s = 6$ dimensional end state space, and were sampled uniformly to cover an area of $750$ cm$^2 \times 3.1$ rad. 

\subsection{Generative model training}
\label{sec:exp:generative_model}
The generative models are represented by neural networks that map a low-dimensional action latent variable $\alpha$ into a $7\times79$ dimensional vector representing $7$ motor actions and $79$ time-steps. In total, we trained $9$ $\beta$-VAE models and $9$ InfoGAN models with latent dimension chosen from $N_\alpha \in \{2, 3, 6\}$.
We refer the reader to the Appendix \ref{app:gen_models} for the exact architecture of the models as well as all the training details. The prior distribution $p(\alpha)$ is the standard normal distribution $\text{N}(0, 1)$ in case of $\beta$-VAEs, and the uniform distribution $\text{U}(-1, 1)$ in case of InfoGANs. 

Table~\ref{tabel:exp:vae} summarizes the parameters of the $\beta$-VAE models together with the values of both the KL divergence and reconstruction term (right and left term in~\eqref{eq:vae}, respectively) obtained at the last epoch.
At the beginning of training, we set $\beta = 0$ and gradually increase its value until the value of the KL divergence drops below a predetermined threshold 
set to $1.5, 2.5$ and $3.5$. The resulting $\beta$ value is reported in Table \ref{tabel:exp:vae} and kept fixed until the end of the training. 
Table \ref{tabel:exp:gan} summarizes the training parameters and loss function values of the InfoGAN models. We report the total model loss~\eqref{eq:gan} (M-loss), the generator loss (G-loss, middle term in~\eqref{eq:gan}) and the mutual information loss (I-loss, right term in~\eqref{eq:gan}) obtained on the last epoch. The hyperparameter $\lambda$ (right term in~\eqref{eq:gan}) was chosen from $\lambda \in \{0.1, 1.5, 3.5\}$.

\begin{table}[H]
    \centering
    \caption{Training parameters of the VAE models together with the values of the loss function \eqref{eq:vae} obtained at the last epoch.}
    \begin{tabular}{c|c|c|c|c}
         \textbf{index} & \textbf{latent size} $N_\alpha$ & $\beta$ & \textbf{KL loss} & \textbf{reconstruction loss}  \\
         \hline
         VAE1 & 2 & $\scnum{1.6e-02}$ & $1.5$ & $\scnum{2.1e-02}$ \\
         VAE2 & 2 & $\scnum{6.4e-03}$ & $2.5$ & $\scnum{1.1e-02}$ \\
         VAE3 & 2 & $\scnum{3.2e-03}$ & $3.4$ & $\scnum{6.7e-03}$ \\
         VAE4 & 3 & $\scnum{1.6e-02}$ & $1.5$ & $\scnum{2.1e-02}$ \\
         VAE5 & 3 & $\scnum{7.2e-03}$ & $2.4$ & $\scnum{1.1e-02}$ \\
         VAE6 & 3 & $\scnum{3.2e-03}$ & $3.5$ & $\scnum{5.5e-03}$ \\
         VAE7 & 6 & $\scnum{1.6e-02}$ & $1.5$ & $\scnum{2.1e-02}$ \\
         VAE8 & 6 & $\scnum{7.2e-03}$ & $2.4$ & $\scnum{1.1e-02}$ \\
         VAE9 & 6 & $\scnum{3.6e-03}$ & $3.3$ & $\scnum{6.0e-03}$ \\
    \end{tabular}
    \label{tabel:exp:vae}
\end{table}

\begin{table}[H]
    \centering
    \caption{Training parameters of the InfoGAN models together with values of the loss function \eqref{eq:gan} obtained at the last epoch. We report the generator loss (G loss), the information loss (I loss) and the total model loss (M loss).}
    \begin{tabular}{c|c|c|c|c|c}
         \textbf{Model} & \textbf{latent size} $N_\alpha$ & $\lambda$ & \textbf{G loss} & \textbf{I loss} & \textbf{M loss} \\
         \hline
         GAN1 & 2 & $0.1$ & $2.81$ & $0.10$ & $0.54$  \\
         GAN2 & 2 & $1.5$ & $2.28$ & $0.59$ & $0.78$ \\
         GAN3 & 2 & $3.5$ & $2.43$ & $1.07$ & $0.63$ \\
         GAN4 & 3 & $0.1$ & $2.52$ & $0.16$ & $0.61$ \\
         GAN5 & 3 & $1.5$ & $2.27$ & $1.45$ & $0.75$ \\
         GAN6 & 3 & $3.5$ & $2.23$ & $3.16$ & $0.68$ \\
         GAN7 & 6 & $0.1$ & $2.50$ & $0.44$ & $0.63$ \\
         GAN8 & 6 & $1.5$ & $2.23$ & $4.78$ & $0.73$ \\
         GAN9 & 6 & $3.5$ & $2.27$ & $10.32$ & $0.67$\\
    \end{tabular}
    \label{tabel:exp:gan}
\end{table}

\subsection{Evaluating the generative models}
\label{sec:exp:eval_gen_models}
We evaluated all the generative models using disentanglement, local linearity, precision and recall measures introduced in Section \ref{sec:eval_generative_model} as well as using the training parameters described in Tables~\ref{tabel:exp:vae} and~\ref{tabel:exp:gan}. We first studied the correlation of each individual evaluation measure and the performance of the policy training, and then the combined effect of all of the measures together. Our analysis empirically shows that any generative model having high recall improves the data-efficiency of the policy training task. Moreover, we observe that $\Dir$ is especially relevant in case of GANs, and precision in case of VAEs, while we do not find local linearity as essential for successful policy training. The details of our evaluation are presented and discussed in detail below.

\begin{figure}[h]
\centering
\includegraphics[width=1.0\linewidth]{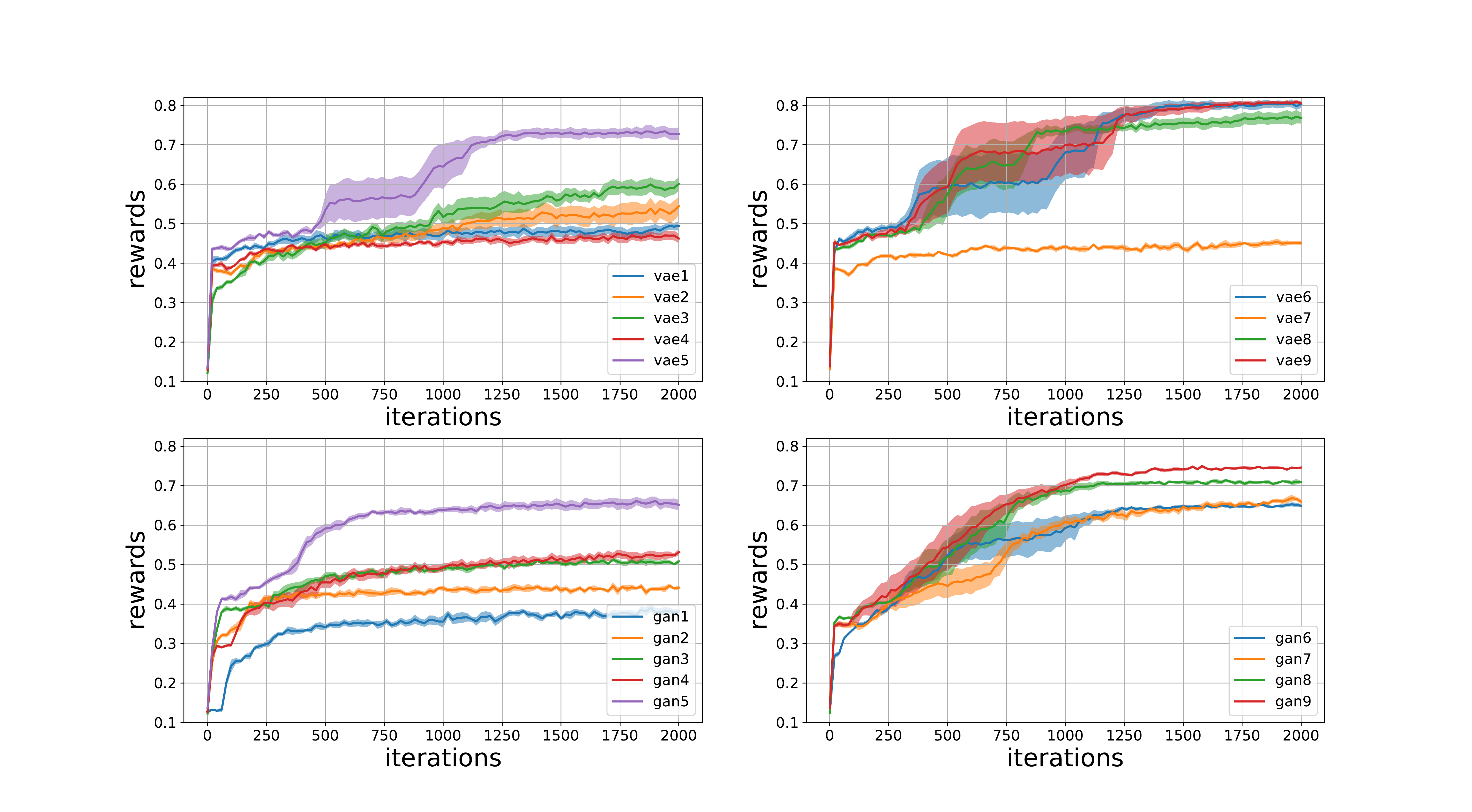}
\caption{A visualization of the performance of the policy training. We report the average reward together with the standard deviation ($y$-axis) obtained during EM policy training ($x$-axis) across three different random seeds. }
\label{fig:exp:policy_training_performance}
\end{figure}

\textbf{EM policy training:}
\label{sec:exp:em_training}
For each generative model, introduced in Tables~\ref{tabel:exp:vae} and~\ref{tabel:exp:gan}, we trained one policy with three different random seeds. The obtained average training performances together with the standard deviations are shown in Figure~\ref{fig:exp:policy_training_performance}.  Using these results, we labeled each generative model with the
maximum reward achieved during EM policy training across all three random seeds. As it can be seen from Figure~\ref{fig:exp:policy_training_performance}, the best performance is achieved with VAE6 and VAE9 which have latent size $N_\alpha$ equal to 3 and 6, respectively, and low $\beta$ values.

We investigated the correlation between the characteristics of the generative models and the performance of the policy training in two different ways: (1) we calculated the Pearson's correlation between each \textit{individual} model characteristic and the corresponding model label introduced above, and (2) we studied the \textit{combined} effect that all of the properties have on the policy training using automatic relevance determination (ARD). We fit one ARD model for VAEs and GANs separately using the evaluation results discussed in the following and the training parameters from Tables \ref{tabel:exp:vae} and \ref{tabel:exp:gan} as inputs, and the policy performance as target values. Since ARD is sensitive to outliers, we additionally preprocessed the input values to the ARD using robust scaling with median and interquartile range. The results including Pearson's correlation coefficients (the higher, the better $\uparrow$) together with the corresponding p-values (the lower, the better $\downarrow$) as well as the ARD estimated precision of weights (the lower, the better $\downarrow$) are shown in Tables~\ref{tab:vae_pear_coef} and~\ref{tab:gan_pear_coef}. We first present the evaluation results and Pearson's correlations of each of the measures separately, and then discuss the differences with the ARD results.

\begin{table}[h]
    \centering
    \caption{Results for the correlation analysis of the VAE models to the policy performance. We report (1) Pearson correlation coefficient R (the higher, the better $\uparrow$) and (2) ARD estimated precision of the weights (the lower, the better $\downarrow$) between the evaluation metrics from Section \ref{sec:eval_generative_model} together with the training parameters from Table~\ref{tabel:exp:vae}, and the policy performance for VAE models visualised in Figure~\ref{fig:exp:policy_training_performance}.} 
    \vspace{0.2cm}
    \label{tab:vae_pear_coef}
    \begin{tabular}{r|c|c|c|c|c|c|c}
    & 
    \textbf{DiP} & \textbf{DiR} &
    \textbf{L3} & \textbf{Precision} & \textbf{Recall} & $\boldsymbol{N_\alpha}$ & $\boldsymbol{\beta}$
    \\
    \hline
    Pearson's R & 
    $0.600$ & 
    \cellcolor[gray]{0.94}{$0.668$} & 
    $-0.100$ & 
    \cellcolor[gray]{0.82}{$0.776$} &  \cellcolor[gray]{0.6}$0.969$ & $0.317$ & 
    \cellcolor[gray]{0.80}{$-0.791$} \\ 
        
    p-value & 
    $0.088$ &  \cellcolor[gray]{0.94}$0.049$ &  
    $0.797$ &  \cellcolor[gray]{0.82}$0.014$ &  \cellcolor[gray]{0.6}$0.000$ &  
    $0.406$ &  \cellcolor[gray]{0.8}$0.011$
    \\
    
    \hline
    ARD &  
    $\scnum{8.07e+04}$ &  $\scnum{1.81e+04}$ &  \cellcolor[gray]{0.8}$\scnum{1.09e+04}$ &  \cellcolor[gray]{0.7}$\scnum{4.45e+03}$ &  \cellcolor[gray]{0.60}$\scnum{1.93e+01}$ &  
    \cellcolor[gray]{0.9}$\scnum{1.70e+04}$ &  $\scnum{1.90e+04}$
    \end{tabular}
\end{table}

\begin{table}[h]
    \centering
    \caption{Results for the correlation analysis of the GAN models to the policy performance. We report (1) Pearson correlation coefficient R (the higher, the better $\uparrow$) and (2) ARD estimated precision of the weights (the lower, the better $\downarrow$) between the evaluation metrics from Section \ref{sec:eval_generative_model} together with the training parameters from Table~\ref{tabel:exp:gan}, and the policy performance for VAE models visualised in Figure~\ref{fig:exp:policy_training_performance}. } 
    \vspace{0.2cm}
    \label{tab:gan_pear_coef}
    \begin{tabular}{r|c|c|c|c|c|c|c}
    & 
    \textbf{DiP} & \textbf{DiR} & \textbf{L3} & \textbf{Precision} & \textbf{Recall} &  $\boldsymbol{N_\alpha}$ & $\boldsymbol{\lambda}$
    \\
    \hline
    Pearson's R & 
    $0.395$ &
    \cellcolor[gray]{0.82}$0.781$ &
    $-0.639$ &
    \cellcolor[gray]{0.80}$-0.801$ &  \cellcolor[gray]{0.65}$0.948$ &    \cellcolor[gray]{0.78}$0.823$ & 
    $0.368$  
    \\
    
    p-value & 
    $0.293$ &  \cellcolor[gray]{0.82}$0.013$ &  
    $0.064$ &  \cellcolor[gray]{0.80}$0.009$ &  \cellcolor[gray]{0.65}$0.000$ &  \cellcolor[gray]{0.78}$0.006$ &  
    $0.330$
    
    \\
    \hline
    ARD & \cellcolor[gray]{0.80}$\scnum{2.37e+02}$ &  \cellcolor[gray]{0.60}$\scnum{5.71e+01}$ &  $\scnum{3.20e+02}$ &  $\scnum{7.74e+02}$ &  \cellcolor[gray]{0.70}$\scnum{1.15e+02}$ &  
    \cellcolor[gray]{0.85}$\scnum{2.67e+02}$ &  $\scnum{5.50e+04}$ 
    \\
         
    \multicolumn{6}{c}{} \\
    &
    \textbf{G loss} & \textbf{I loss}&      \multicolumn{1}{c}{\textbf{M loss}}
    
    \\
    \cline{1-4}
    Pearson's R & 
    $-0.672$ & 
    $0.698$ & \multicolumn{1}{c}{$0.381$}
    
    \\
    p-value & 
    $0.048$ & 
    $0.036$ &  \multicolumn{1}{c}{$0.312$} 
    \\
    
    \cline{1-4}
    ARD & 
    $\scnum{4.66e+03}$ &  $\scnum{1.19e+05}$ &   \multicolumn{1}{c}{$\scnum{3.56e+02}$}
    \end{tabular}
\end{table}

\begin{figure}[h]
\centering
\includegraphics[width=0.6\linewidth]{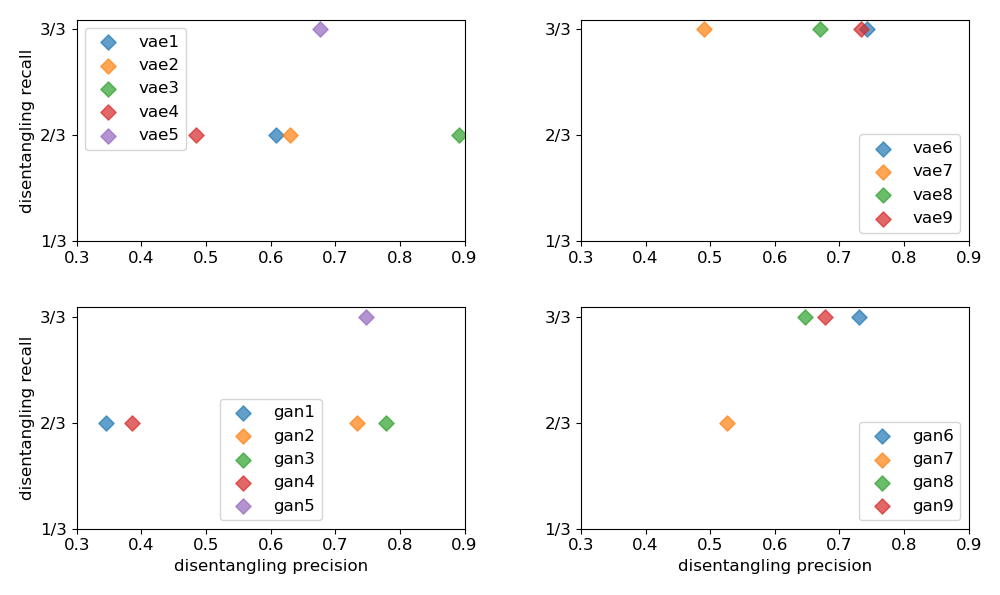}
\caption{Disentangling precision and recall scores for VAE models (top row) and GAN models (bottom row) with training parameters reported in Tables~\ref{tabel:exp:vae} and \ref{tabel:exp:gan}, respectively.} \label{fig:dis_all}
\end{figure}

\textbf{Disentanglement:}
We measured the disentangling precision and recall using $\MMD$ with kernel parameter $\gamma = 15$. For a given model $g$, we performed $D = 5$ interventions on every latent dimension $l \in \{1, \dots, N_\alpha\}$. 
We restricted the dimension of the end state space to $N_s = 3$ including only the position of the object and the angle of picking since these are the most relevant factors for performing the picking task.
The size of the sets $\boldsymbol{S}_g^{l-I_d}$ and $\boldsymbol{S}_r$ containing the end states was set to $n = 200$. For complete evaluation details we refer the reader to Appendix \ref{app:dis_score}. The obtained disentanglement scores are visualised in Figure \ref{fig:dis_all}. We observe that models with higher latent space dimension $N_\alpha$ (Figure \ref{fig:dis_all}, right) on average achieve higher disentangling recall $\Dir$. In Tables~\ref{tab:vae_pear_coef} and~\ref{tab:gan_pear_coef} we see that $\Dir$ is positively correlated to the performance of the policy, with a stronger correlation in case of GANs, while the positive correlation of $\Dip$ is not significant (p-value $> 0.05$). Therefore, when choosing  a generative model based only on disentanglement, it is beneficial that the interventions on the latent dimensions affect all three dimensions of the end state space. In other words, the latent action representations should capture all aspects of the robotic task, while it is less important to capture them precisely. 

\textbf{Local linearity:}
Given a generative model $g$, we randomly sampled $50$ latent actions from the prior $p(\alpha)$. For each such latent action $\alpha$, we set $\varepsilon = 0.2$ and sampled $500$ points from its $\varepsilon$-neighbourhood $N_\varepsilon(\alpha)$. We then fit an affine transformation $f_\alpha$ defined in~\eqref{eq:aff_trans} on $350$ randomly selected neighbourhood points and calculated L3 as the test MSE on the remaining $150$ points. We report the average test MSE obtained across all $50$ points in Table~\ref{tabel:exp:loclin}. We observe that all VAEs except for VAE3 achieve a lower test MSE than GAN models. By comparing Figure~\ref{fig:exp:policy_training_performance} and Table~\ref{tabel:exp:loclin} it appears that local linearity is not related to the performance of the policy training which is on par with the results obtained in 
Tables~\ref{tab:vae_pear_coef} and~\ref{tab:gan_pear_coef} where we see that the correlation is not significant neither for VAEs nor for GANs. Despite being insignificant, the correlation is negative for all the models which coincides with our hypothesis that a more locally linear model (i.e., a model with a lower MSE) performs better in the policy training.
 

\begin{table}[H]
    \centering
    \caption{The Latent Local Linearity (L3) results measured as the mean squared error (MSE) of the affine transformations defined in~\eqref{eq:aff_trans} in the neighbourhoods of 50 latent action representations for all VAE and GAN models.}
    \vspace{0.2cm}
    \begin{tabular}{c c c c c c c c c c}
         \textbf{Model} & VAE1 & VAE2 & VAE3 & VAE4 & VAE5 & VAE6 & VAE7 & VAE8 & VAE9  \\
         \hline
         \textbf{MSE} & 1.6& 3.2& 14.3& 1.5& 1.6& 2.0& 1.2& 1.2& 1.3\\
         & &  &  &  &  &  &  &  & \\
         \textbf{Model}&GAN1 & GAN2 & GAN3 &GAN4 & GAN5 & GAN6 &GAN7 & GAN8 & GAN9\\
         \hline
         \textbf{MSE} &73.1&27.8&41.1&16.7&36.6&43.0&14.2&24.7&5.2 \\
    \end{tabular}
    \label{tabel:exp:loclin}
\end{table}

\begin{figure}[h]
\centering
\includegraphics[width=0.6\linewidth]{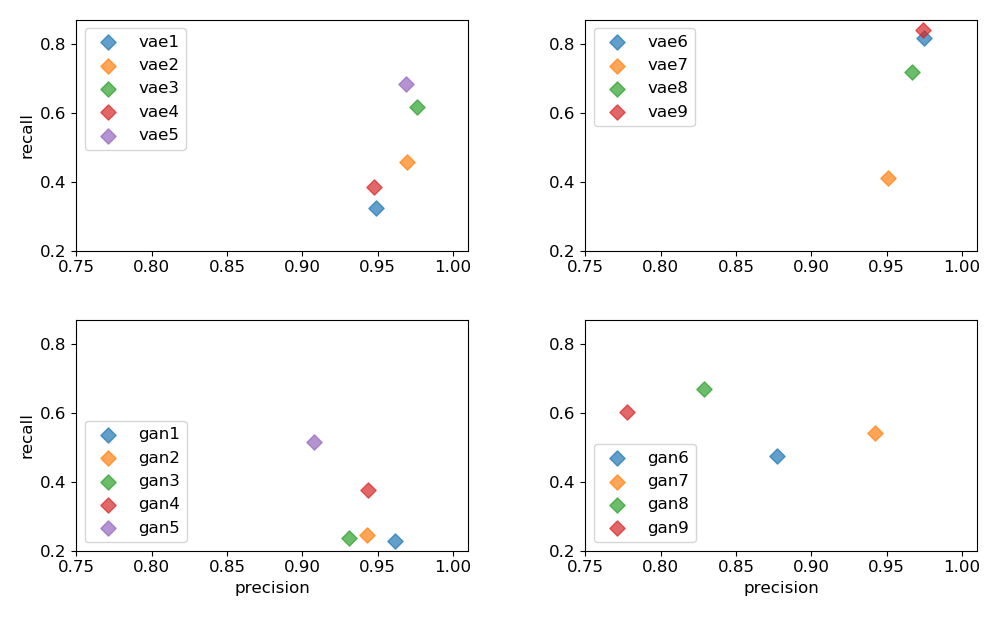} 
\caption{Precision and recall scores for VAE (top row) and GAN models (bottom row).}\label{fig:ipr_all}
\end{figure}

\textbf{Precision and recall:}
For each generative model $g$, we randomly sampled $15000$ samples from the latent prior distribution $p(\alpha)$. The corresponding set of the generated trajectories $\boldsymbol{T_g}$ was compared to a set $\boldsymbol{T_r}$ of $15000$ randomly chosen training trajectories  which were sampled only once and fixed for all the models. The neighbourhood size $k$ was set to $3$ as suggested in \cite{kynkaanniemi2019improved}. The resulting precision and recall scores are shown in Figure \ref{fig:ipr_all}. Firstly, we observe that all the models have relatively high precision except for GAN8-9 which are shown in the bottom right part of Figure~\ref{fig:ipr_all}. Secondly, on average GANs have worse recall than VAEs, which is a consequence of the training procedure (see Appendix \ref{app:gen_models}) and can possibly be improved with a more thorough fine tuning of the models. This is consistent with the results reported in Table~\ref{tab:gan_pear_coef} where precision is negatively correlated to the policy performance in case of GANs but positively in case of VAEs. For both VAE and GAN models, we observe that recall is highly correlated with the performance of the policy training.

 
\textbf{Training parameters:} In Tables~\ref{tab:vae_pear_coef} and~\ref{tab:gan_pear_coef} we additionally calculated the Pearson's correlation between the training parameters, shown in Tables~\ref{tabel:exp:vae} and~\ref{tabel:exp:gan}, and the policy performance shown in Figure~\ref{fig:exp:policy_training_performance}. For VAEs, we used only the $\beta$ coefficient as it is clearly correlated with the KL divergence and the reconstruction loss (see Table~\ref{tabel:exp:vae}). Interestingly, we observe negative correlation between $\beta$ and policy training performance which indicates that  disentangled representations are less beneficial as these are obtained with a higher $\beta$ value. This observation is also supported by a low Pearson's coefficient obtained for $\Dir$ and insignificant correlation of $\Dip$ (see Table~\ref{tab:vae_pear_coef}). In fact, increasing $\beta$ positively affects the disentanglement as shown in~\cite{higgins2017beta} on image data, but negatively the precision, because data is in this case reconstructed from a wider approximate posterior distribution $q_\varphi$. Since precision is more important for an efficient policy performance than disentanglement in case of VAEs, the correlation of $\beta$ is negative.
Next, a positive correlation of the latent dimension $N_\alpha$ is observed for both VAE and GAN models but for VAEs it is insignificant. Finally, the correlation coefficients for GAN losses imply that the lower the generator loss (G loss) and the higher the information loss (I loss), the better the policy performance. This result is meaningful as higher I loss, obtained with a higher $\lambda$, encourages more disentangled representations, which we also see with a high $\Dir$ correlation coefficient. 


\textbf{Correlation between evaluation metrics and EM policy training:} Possibly due to the sensitivity of the ARD framework to outliers, the obtained ARD results shown in Tables~\ref{tab:vae_pear_coef} and \ref{tab:gan_pear_coef} differ slightly from the Pearson's R correlation coefficients. In case of VAEs, local linearity seems to be more important due to the large outlying value for the VAE3 model (Table~\ref{tabel:exp:loclin}). Similarly, $\Dip$ appears to be influential in case of GANs due to the large range of values (Figure
~\ref{fig:dis_all} left). For both model types, ARD analysis shows $N_\alpha$ to be important which can also be a consequence of the chosen values $N_\alpha \in \{2, 3, 6\}$. However, ARD scores precision and recall as influential in case of VAEs, as well as $\Dir$ and recall in case of GANs, both of which matches the Pearson's correlation results.

Lastly, we wished to compare all the generative models only based on the properties that can be evaluated for both model types. Therefore, we fit three ARD models, one for VAEs, one for GANs and one for all the models combined using only $\Dip$, $\Dir$, local linearity, precision and recall. The obtained estimated precision of the weights are reported in Table~\ref{tab:ard_coef}. While we observe minor differences between the two model types, we see that the most important property of a generative model combined with EM policy training is recall which is consistent with the results obtained in Tables
~\ref{tab:vae_pear_coef} and \ref{tab:gan_pear_coef}. Similarly as before, disentangling precision $\Dip$ gains importance due to the large range of values (Figure~\ref{fig:dis_all}). Therefore, for successful policy training in case of the picking task, it is crucial that the generative model is able to reproduce samples from the training dataset even if the generated samples are less precise. Based on the observations from Tables
~\ref{tab:vae_pear_coef} and \ref{tab:gan_pear_coef} we conclude that GAN models additionally benefit from a higher disentangling recall $\Dir$, while VAEs benefit from a higher precision. Finally, we note that disentangling precision as well as local linearity can become more important when performing tasks represented by more complex data where structured latent representations would be beneficial.


\begin{table}[h]
    \centering
    \caption{ARD estimated precision of the weights (the lower, the better $\downarrow$) corresponding to the evaluation metrics from Section \ref{sec:eval_generative_model} for all the VAE and GAN models.}     
    \vspace{0.2cm}
    \label{tab:ard_coef}
    \begin{tabular}{r|c|c|c|c|c}
        \textbf{models} 
          & \textbf{DiP} & \textbf{DiR}
          & \textbf{L3}  
          & \textbf{Precision} & \textbf{Recall}   \\
         \hline
          VAEs & \cellcolor[gray]{0.75}$\scnum{2.78e+02}$ &  
          $\scnum{1.83e+04}$ 
          &  
          \cellcolor[gray]{0.85}$\scnum{9.42e+03}$ &  
          $\scnum{2.24e+04}$ 
          &  
          \cellcolor[gray]{0.65}$\scnum{3.56e+01}$
         \\
         \hline
         GANs & \cellcolor[gray]{0.75}$\scnum{8.17e+02}$ &  
         \cellcolor[gray]{0.85}$\scnum{1.09e+04}$ &  
         $\scnum{1.25e+04}$ &  $\scnum{3.97e+04}$ &  
         \cellcolor[gray]{0.65}$\scnum{2.66e+01}$ \\
         \hline
         
         ALL & \cellcolor[gray]{0.75}$\scnum{3.62e+03}$ &  
         $\scnum{1.60e+04}$ & $\scnum{1.51e+04}$ 
         &  
         \cellcolor[gray]{0.85}$\scnum{6.27e+03}$ & 
         \cellcolor[gray]{0.65}$\scnum{3.45e+01}$
          \\

    \end{tabular}
\end{table}

%% file: inputs/conclusion.tex
\section{Conclusion}
\label{sec:conclusion}
We presented an RL framework that combined with generative models trains deep visuomotor policies in a data-efficient manner. The generative models are integrated with the RL optimization by introducing a latent variable $\alpha$ that is a low-dimensional representation of motor actions. Using the latent action variable $\alpha$, we divided the optimization of the parameters $\Theta$ of the deep visuomotor policy $\pi_\Theta(\tau|s)$
into two parts: optimizing the parameters $\vartheta$ of a generative model $p_\vartheta(\tau|\alpha)$ that generates valid sequences of motor actions, and optimizing the parameters $\theta$ of a sub-policy $\pi_\theta(\alpha | s)$, where $\Theta = [\theta, \vartheta]$. The sub-policy parameters $\theta$ are found using the EM algorithm, while generative model parameters $\vartheta$ are trained unsupervised to optimize the objective corresponding to the chosen generative model. In summary, the complete framework consists of three data-efficient downstream tasks: 
(a) training the generative model $p_\vartheta$, (b) training the sub-policy $\pi_\theta$, and (c) supervised end-to-end training the deep visuomotor policy $\pi_\Theta$.

Moreover, we provided a set of measures for evaluating the quality of the generative models regulated by the RL policy search algorithms such that we can predict the performance of the deep policy training $\pi_\Theta$ prior to the actual training. In particular, we defined two new measures,  disentangling precision and recall ($\Dip$ and $\Dir$) and latent local linearity (L3), that evaluate the quality of the latent space of the generative model $p_\vartheta$, and complemented them with precision and recall measure \cite{kynkaanniemi2019improved} which evaluates the quality of the generated samples. We experimentally demonstrated the predictive power of these measures on a picking task using a set of different VAE and GAN generative models. 
Regardless of the model type, we observe recall to be the most influential property, followed by precision in case of VAEs and by disentangling recall in case of GANs.

%% file: inputs/appendix.tex
\begin{appendices}

\input{inputs/perception}

\section{Disentangling Precision and Recall} \label{app:dis_score}
Let $l \in \{1, \dots, N_\alpha\}$ be a fixed latent dimension. In phase $1$, we perform $D \in \mathbb{N}$ interventions $\alpha_l = I_d$ on the latent dimension $l$ where $d = 1, \dots, D$. Interventions are chosen from the set of equidistant points on the interval $[-a, a]$ such that $I_d = -a + 2a \cdot \frac{d-1}{D - 1}$. The value $a$ is chosen such that $[-a, a]$ is in the support of the prior distribution $p(\alpha)$. Since $p(\alpha)$ is a standard normal distribution in the case of VAEs and a uniform distribution on the interval $[-1, 1]$ in the case of GANs, we set $a$ to be $1.5$ and $1$ in the case of the VAEs and GANs, respectively. Each intervention $d = 1, \dots, D$ is performed on $n$ samples from the prior distribution $p(\alpha)$ and yields a set of $n$ end states  denoted by $\boldsymbol{S}_g^{l-I_d}$. For each intervention we additionally randomly sample a set $\boldsymbol{S}_r$ of $n$ end states corresponding to the training motor data. Note that elements of both $\boldsymbol{S}_g^{l-I_d}$ and $\boldsymbol{S}_r$ are $N_s$-dimensional with $N_s$ being the dimension of the end state space, which are obtained by executing the generated trajectories on the robotic platform..

In phase $2$, we calculate the $\MMD(\proj_j \boldsymbol{S}_g^{l-I_d}, \proj_j \boldsymbol{S}_r)$ for a fixed intervention $d = 1, \dots, D$ and every end state component $j = 1, \dots, N_s$. We first determine if the difference between the sets $\proj_j \boldsymbol{S}_g^{l-I_d}$ and $\proj_j \boldsymbol{S}_r$ is large enough to reject the null hypothesis that samples from $\proj_j \boldsymbol{S}_g^{l-I_d}$ and $\proj_j \boldsymbol{S}_r$ are drawn from the same distribution. We achieve this by performing a permutation test where we pool all the samples from $\proj_j \boldsymbol{S}_g^{l-I_d}$ and $\proj_j \boldsymbol{S}_r$, randomly divide the pooled set into two sets of $n$ elements and calculate the $\MMD$ between the obtained sets. The random split into two sets is performed $100$ times such that we obtain a distribution over the resulting $\MMD$ values. For a predetermined significance level $\eta$, we define the critical value $c_\eta$ to be $(1 - \eta)$-quantile of the obtained distribution over $\MMD$ values.
We then say that the intervention $I_d$ was significant for an end state component $j$ if the observed $\MMD(\proj_j \boldsymbol{S}_g^{l-I_d}, \proj_j \boldsymbol{S}_r) > c_\eta$. The calculations in phase $2$ were repeated $p$ times with a resampled set of $n$ training end states $\boldsymbol{S}_r$. In all our experiments we set $p = 10$ and $\eta = 0.001$. 

Therefore, phases $1$ and $2$ yield functions $c_g: \{1, \dots, N_\alpha\} \longrightarrow \{1, \dots, N_s\}$ and $d_g: \{1, \dots, N_\alpha\} \longrightarrow \mathbb{R}$ defined by:
\begin{align*}
    c_g(l) = \argmax_{j = 1, \dots, N_s} \overline{\MMD}\left(\proj_j \boldsymbol{S}_g^{l-I_d}, \proj_j \boldsymbol{S}_r\right) \quad \textrm{and} \quad d_g(l) = \overline{\MMD}\left(\proj_{c_g(l)} \boldsymbol{S}_g^{l-I_d},  \proj_{c_g(l)} \boldsymbol{S}_r\right)
\end{align*}
where $\overline{\MMD}$ denotes the average $\MMD$ score calculated on a subset of $p \cdot D$ performed interventions that were significant. For a given latent dimension $l \in \{1, \dots, N_\alpha\}$, $c_g(l)$ represents the dimension of the end state space $\mathbb{R}^{N_s}$ that was most affected by the latent interventions. This is because a high $\MMD$ value indicates a low similarity between $\proj_j \boldsymbol{S}_g^{l-I_d}$ and $\proj_j \boldsymbol{S}_r$, and thus a high effect of the intervention. Moreover, $d_g(l)$ is the average $\MMD$ value obtained on the most affected end state space dimension identified by $c_g(l)$.

In phase $3$ we define the final disentanglement score for the generative model $g$ using the functions $c_g$ and $d_g$. Let $\mathcal{P}$ be a subset of $\{ d_g(l): l = 1 \dots, N_\alpha\}$ containing its largest three elements, i.e., the three largest $\overline{\MMD}$ values obtained in phase 2, and let $\mathcal{R} = \{c_g(l): d_g(l) \in \mathcal{P}\}$ be the set of the corresponding end state components. We define the \textit{Disentangling Precision and Recall} $\Dis$ as a pair
\begin{align}
    \Dis(g) = (\Dip(g), \Dir(g)) := \left(\sum_{d \in \mathcal{P}} d, \frac{|\mathcal{R}^{\neq}|}{N_s} \right)
\end{align}
where $\mathcal{R}^{\neq}$ denotes the subset of unique elements of the set $\mathcal{R}$. The \textit{disentangling recall} $\Dir$ is the fraction of end state dimensions described by three most significant latent dimensions, i.e., by the three latent dimensions on which interventions yielded the largest changes in the end state space. The larger the $\Dir$ value, the more end state space dimensions are captured in the latent space, and thus the latent disentanglement has a higher recall.
Similarly, the \textit{disentangling precision} $\Dip$ is the sum effect that the latent interventions on the three most significant latent dimensions have on the affected end state dimensions. The larger the $\Dip$, the stronger was the effect of the latent interventions, and thus the latent disentanglement is more precise.

\section{Generative models} \label{app:gen_models}
\subsection{Variational Autoencoder}
The architecture of the decoder neural network is visualised in Table \ref{tab:gen_arc}. The encoder neural network is symmetric to the decoder with two output linear layers of size $N_\alpha$ representing the mean and the log standard deviation of the approximate posterior distribution. All the models were trained for $10000$ epochs with learning rate fixed to $1e-4$.

\subsection{InfoGAN}
The architecture of the generator, discriminator and Q neural network parametrizing $Q_\phi(\alpha|\tau)$ are summarised in Tables \ref{tab:gen_arc} and \ref{tab:dis_arc}. All the models were trained for $1000$ epochs with learning rates of the optimizers for the generator and discriminator networks fixed to $2e-4$. 

\begin{table}[!htb]
    \begin{minipage}{.4\linewidth}
      \caption{Architecture of the generator neural network.} \label{tab:gen_arc}
      \centering
        \begin{tabular}{l}
         \hline
         \hline
         Linear($N_\alpha$, $128$) + BatchNorm + ReLU \\ 
         \hline
         Linear($128$, $256$) + BatchNorm + ReLU \\
         \hline
         Linear($256$, $512$) + BatchNorm + ReLU \\
         \hline
         Linear($512$, $7 \cdot 79$)
    \end{tabular}
    \end{minipage}%
    \hfill
    \begin{minipage}{.4\linewidth}
      \centering
        \caption{Architecture of the discriminator and Qnet neural networks.} \label{tab:dis_arc}
        \begin{tabular}{l|l}
         \hline
         \hline
         \multirow{2}{*}{Shared layers} & Linear($7 \cdot 79$, $256$) + ReLU \\ 
         \cline{2-2}
         & Linear($256$, $128$) + ReLU \\
         \hline
         discriminator & Linear($128$, $1$) + Sigmoid \\
         \hline \multirow{2}{*}{Qnet} & Linear($128$, $64$) \\ \cline{2-2}
          & Linear($64$, $N_\alpha$) 
         
         \end{tabular}
    \end{minipage} 
\end{table}

\end{appendices}

%% file: inputs/perception.tex
\section{End-to-end Training of Perception and Control}
\label{sec:perception}
The EM policy training algorithm presented in Section~\ref{sec:em_policy_training} updates the deep policy using the supervised learning objective function introduced in~\eqref{eq:M_loss} (the M-step objective). 
Similar to GPS \cite{levine2016end}, the EM policy training formulation enables simultaneous training of the perception and control parts of the deep policy in an end-to-end fashion. 
In this section, we describe two techniques that can improve the efficiency of the end-to-end training.

\textbf{Input remapping trick}
The input remapping trick \cite{levine2016end} can be applied to condition the variational policy $q$ on a low-dimensional compact state representation, $z$, instead of the high-dimensional states $s$ given by the sensory observations, e.g., camera images. 
The policy training phase can be done in a controlled environment such that extra measures other than the sensory observation of the system can be provided. These extra measures can be for example the position of a target object on a tabletop. 
Therefore, the image observations $s$ can be paired with a compact task-specific state representation $z$ such that $z$ is used in the E-step for updating the variational policy $q_\phi(\alpha|z)$, and $s$ in the M-step for updating the policy $\pi_\theta(\alpha|s)$. 

\textbf{Domain adaptation for perception training}
Domain adaptation techniques, e.g., adversarial methods \cite{chen2019adversarial}, can improve the end-to-end training of visuomotor policies with limited robot data samples. 
The unlabeled task-specific images, captured without involving the robot, can be exploited in the M-step to improve the generality of the visuomotor policy to manipulate novel task objects in cluttered backgrounds. 

The M-step is updated to include an extra loss function to adapt data from the two different domains: (i) unlabeled images and (ii) robot visuomotor data. 
The images must contain only one task object in a cluttered background, possibly different than the task object used by the robot during the policy training. 
Given images from the two domains, the basic idea is to extract visual features such that it is not possible to detect the source of the features. 
More details of the method can be found in our recent work in \cite{chen2019adversarial}.

%% file: main.bbl
\begin{thebibliography}{10}
\providecommand{\url}[1]{#1}
\csname url@samestyle\endcsname
\providecommand{\newblock}{\relax}
\providecommand{\bibinfo}[2]{#2}
\providecommand{\BIBentrySTDinterwordspacing}{\spaceskip=0pt\relax}
\providecommand{\BIBentryALTinterwordstretchfactor}{4}
\providecommand{\BIBentryALTinterwordspacing}{\spaceskip=\fontdimen2\font plus
\BIBentryALTinterwordstretchfactor\fontdimen3\font minus
  \fontdimen4\font\relax}
\providecommand{\BIBforeignlanguage}[2]{{%
\expandafter\ifx\csname l@#1\endcsname\relax
\typeout{** WARNING: IEEEtran.bst: No hyphenation pattern has been}%
\typeout{** loaded for the language `#1'. Using the pattern for}%
\typeout{** the default language instead.}%
\else
\language=\csname l@#1\endcsname
\fi
#2}}
\providecommand{\BIBdecl}{\relax}
\BIBdecl

\bibitem{ghadirzadeh2017deep}
A.~Ghadirzadeh, A.~Maki, D.~Kragic, and M.~Bj{\"o}rkman, ``Deep predictive
  policy training using reinforcement learning,'' in \emph{2017 IEEE/RSJ
  International Conference on Intelligent Robots and Systems (IROS)}.\hskip 1em
  plus 0.5em minus 0.4em\relax IEEE, 2017, pp. 2351--2358.

\bibitem{arndt2019meta}
K.~Arndt, M.~Hazara, A.~Ghadirzadeh, and V.~Kyrki, ``Meta reinforcement
  learning for sim-to-real domain adaptation,'' in \emph{2020 IEEE
  International Conference on Robotics and Automation (ICRA)}, 2020.

\bibitem{levine2016end}
S.~Levine, C.~Finn, T.~Darrell, and P.~Abbeel, ``End-to-end training of deep
  visuomotor policies,'' \emph{The Journal of Machine Learning Research},
  vol.~17, no.~1, pp. 1334--1373, 2016.

\bibitem{levine2018learning}
S.~Levine, P.~Pastor, A.~Krizhevsky, J.~Ibarz, and D.~Quillen, ``Learning
  hand-eye coordination for robotic grasping with deep learning and large-scale
  data collection,'' \emph{The International Journal of Robotics Research},
  vol.~37, no. 4-5, pp. 421--436, 2018.

\bibitem{kynkaanniemi2019improved}
T.~Kynk{\"a}{\"a}nniemi, T.~Karras, S.~Laine, J.~Lehtinen, and T.~Aila,
  ``Improved precision and recall metric for assessing generative models,'' in
  \emph{Advances in Neural Information Processing Systems}, 2019, pp.
  3927--3936.

\bibitem{higgins2017beta}
I.~Higgins, L.~Matthey, A.~Pal, C.~Burgess, X.~Glorot, M.~Botvinick,
  S.~Mohamed, and A.~Lerchner, ``beta-vae: Learning basic visual concepts with
  a constrained variational framework,'' in \emph{International Conference on
  Learning Representations}, 2017.

\bibitem{chen2016infogan}
X.~Chen, Y.~Duan, R.~Houthooft, J.~Schulman, I.~Sutskever, and P.~Abbeel,
  ``Infogan: Interpretable representation learning by information maximizing
  generative adversarial nets,'' in \emph{Advances in neural information
  processing systems}, 2016, pp. 2172--2180.

\bibitem{chen2019adversarial}
X.~Chen, A.~Ghadirzadeh, M.~Bj{\"o}rkman, and P.~Jensfelt, ``Adversarial
  feature training for generalizable robotic visuomotor control,'' in
  \emph{2020 IEEE International Conference on Robotics and Automation (ICRA)},
  2020.

\bibitem{hamalainen2019affordance}
A.~H{\"a}m{\"a}l{\"a}inen, K.~Arndt, A.~Ghadirzadeh, and V.~Kyrki, ``Affordance
  learning for end-to-end visuomotor robot control,'' in \emph{2019 IEEE/RSJ
  international conference on intelligent robots and systems (IROS)}, 2019.

\bibitem{butepage2019imitating}
J.~B{\"u}tepage, A.~Ghadirzadeh, {\"O}.~{\"O}. Karadag, M.~Bj{\"o}rkman, and
  D.~Kragic, ``Imitating by generating: deep generative models for imitation of
  interactive tasks,'' \emph{Frontiers in Robotics and AI}, 2020.

\bibitem{finn2016deep}
C.~Finn, X.~Y. Tan, Y.~Duan, T.~Darrell, S.~Levine, and P.~Abbeel, ``Deep
  spatial autoencoders for visuomotor learning,'' in \emph{2016 IEEE
  International Conference on Robotics and Automation (ICRA)}.\hskip 1em plus
  0.5em minus 0.4em\relax IEEE, 2016, pp. 512--519.

\bibitem{kalashnikov2018qt}
D.~Kalashnikov, A.~Irpan, P.~Pastor, J.~Ibarz, A.~Herzog, E.~Jang, D.~Quillen,
  E.~Holly, M.~Kalakrishnan, V.~Vanhoucke, and S.~Levine, ``Qt-opt: Scalable
  deep reinforcement learning for vision-based robotic manipulation,'' in
  \emph{2nd Conference on Robot Learning (CoRL)}, 2018.

\bibitem{quillen2018deep}
D.~Quillen, E.~Jang, O.~Nachum, C.~Finn, J.~Ibarz, and S.~Levine, ``Deep
  reinforcement learning for vision-based robotic grasping: A simulated
  comparative evaluation of off-policy methods,'' in \emph{2018 IEEE
  International Conference on Robotics and Automation (ICRA)}.\hskip 1em plus
  0.5em minus 0.4em\relax IEEE, 2018, pp. 6284--6291.

\bibitem{singh2017gplac}
A.~Singh, L.~Yang, and S.~Levine, ``Gplac: Generalizing vision-based robotic
  skills using weakly labeled images,'' in \emph{Proceedings of the IEEE
  International Conference on Computer Vision}, 2017, pp. 5851--5860.

\bibitem{devin2018deep}
C.~Devin, P.~Abbeel, T.~Darrell, and S.~Levine, ``Deep object-centric
  representations for generalizable robot learning,'' in \emph{2018 IEEE
  International Conference on Robotics and Automation (ICRA)}.\hskip 1em plus
  0.5em minus 0.4em\relax IEEE, 2018, pp. 7111--7118.

\bibitem{pinto2017asymmetric}
L.~Pinto, M.~Andrychowicz, P.~Welinder, W.~Zaremba, and P.~Abbeel, ``Asymmetric
  actor critic for image-based robot learning,'' \emph{arXiv preprint
  arXiv:1710.06542}, 2017.

\bibitem{finn2017deep}
C.~Finn and S.~Levine, ``Deep visual foresight for planning robot motion,'' in
  \emph{2017 IEEE International Conference on Robotics and Automation
  (ICRA)}.\hskip 1em plus 0.5em minus 0.4em\relax IEEE, 2017, pp. 2786--2793.

\bibitem{gu2017deep}
S.~Gu, E.~Holly, T.~Lillicrap, and S.~Levine, ``Deep reinforcement learning for
  robotic manipulation with asynchronous off-policy updates,'' in \emph{2017
  IEEE international conference on robotics and automation (ICRA)}.\hskip 1em
  plus 0.5em minus 0.4em\relax IEEE, 2017, pp. 3389--3396.

\bibitem{dasari2019robonet}
S.~Dasari, F.~Ebert, S.~Tian, S.~Nair, B.~Bucher, K.~Schmeckpeper, S.~Singh,
  S.~Levine, and C.~Finn, ``Robonet: Large-scale multi-robot learning,''
  \emph{arXiv preprint arXiv:1910.11215}, 2019.

\bibitem{abdolmaleki2020distributional}
A.~Abdolmaleki, S.~H. Huang, L.~Hasenclever, M.~Neunert, H.~F. Song,
  M.~Zambelli, M.~F. Martins, N.~Heess, R.~Hadsell, and M.~Riedmiller, ``A
  distributional view on multi-objective policy optimization,'' \emph{arXiv
  preprint arXiv:2005.07513}, 2020.

\bibitem{peng2018sim}
X.~B. Peng, M.~Andrychowicz, W.~Zaremba, and P.~Abbeel, ``Sim-to-real transfer
  of robotic control with dynamics randomization,'' in \emph{2018 IEEE
  international conference on robotics and automation (ICRA)}.\hskip 1em plus
  0.5em minus 0.4em\relax IEEE, 2018, pp. 1--8.

\bibitem{tobin2017domain}
J.~Tobin, R.~Fong, A.~Ray, J.~Schneider, W.~Zaremba, and P.~Abbeel, ``Domain
  randomization for transferring deep neural networks from simulation to the
  real world,'' in \emph{2017 IEEE/RSJ international conference on intelligent
  robots and systems (IROS)}.\hskip 1em plus 0.5em minus 0.4em\relax IEEE,
  2017, pp. 23--30.

\bibitem{yu2018one}
T.~Yu, C.~Finn, A.~Xie, S.~Dasari, T.~Zhang, P.~Abbeel, and S.~Levine,
  ``One-shot imitation from observing humans via domain-adaptive
  meta-learning,'' \emph{arXiv preprint arXiv:1802.01557}, 2018.

\bibitem{chen2018deep}
X.~Chen, A.~Ghadirzadeh, J.~Folkesson, M.~Bj{\"o}rkman, and P.~Jensfelt, ``Deep
  reinforcement learning to acquire navigation skills for wheel-legged robots
  in complex environments,'' in \emph{2018 IEEE/RSJ International Conference on
  Intelligent Robots and Systems (IROS)}.\hskip 1em plus 0.5em minus
  0.4em\relax IEEE, 2018, pp. 3110--3116.

\bibitem{tzeng2017adversarial}
E.~Tzeng, J.~Hoffman, K.~Saenko, and T.~Darrell, ``Adversarial discriminative
  domain adaptation,'' in \emph{Proceedings of the IEEE Conference on Computer
  Vision and Pattern Recognition}, 2017, pp. 7167--7176.

\bibitem{tzeng2020adapting}
E.~Tzeng, C.~Devin, J.~Hoffman, C.~Finn, P.~Abbeel, S.~Levine, K.~Saenko, and
  T.~Darrell, ``Adapting deep visuomotor representations with weak pairwise
  constraints,'' in \emph{Algorithmic Foundations of Robotics XII}.\hskip 1em
  plus 0.5em minus 0.4em\relax Springer, 2020, pp. 688--703.

\bibitem{schulman2015trust}
J.~Schulman, S.~Levine, P.~Abbeel, M.~Jordan, and P.~Moritz, ``Trust region
  policy optimization,'' in \emph{International conference on machine
  learning}, 2015, pp. 1889--1897.

\bibitem{schulman2017proximal}
J.~Schulman, F.~Wolski, P.~Dhariwal, A.~Radford, and O.~Klimov, ``Proximal
  policy optimization algorithms,'' \emph{arXiv preprint arXiv:1707.06347},
  2017.

\bibitem{neumann2011variational}
G.~Neumann, ``Variational inference for policy search in changing situations,''
  in \emph{Proceedings of the 28th International Conference on Machine
  Learning, ICML 2011}, 2011, pp. 817--824.

\bibitem{deisenroth2013survey}
M.~P. Deisenroth, G.~Neumann, J.~Peters \emph{et~al.}, ``A survey on policy
  search for robotics,'' \emph{Foundations and Trends{\textregistered} in
  Robotics}, vol.~2, no. 1--2, pp. 1--142, 2013.

\bibitem{levine2013variational}
S.~Levine and V.~Koltun, ``Variational policy search via trajectory
  optimization,'' in \emph{Advances in neural information processing systems},
  2013, pp. 207--215.

\bibitem{ghadirzadeh2018sensorimotor}
A.~Ghadirzadeh, ``Sensorimotor robot policy training using reinforcement
  learning,'' Ph.D. dissertation, KTH Royal Institute of Technology, 2018.

\bibitem{peters2006policy}
J.~Peters and S.~Schaal, ``Policy gradient methods for robotics,'' in
  \emph{2006 IEEE/RSJ International Conference on Intelligent Robots and
  Systems}.\hskip 1em plus 0.5em minus 0.4em\relax IEEE, 2006, pp. 2219--2225.

\bibitem{peters2008reinforcement}
------, ``Reinforcement learning of motor skills with policy gradients,''
  \emph{Neural networks}, vol.~21, no.~4, pp. 682--697, 2008.

\bibitem{ijspeert2003learning}
A.~J. Ijspeert, J.~Nakanishi, and S.~Schaal, ``Learning attractor landscapes
  for learning motor primitives,'' in \emph{Advances in neural information
  processing systems}, 2003, pp. 1547--1554.

\bibitem{ijspeert2013dynamical}
A.~J. Ijspeert, J.~Nakanishi, H.~Hoffmann, P.~Pastor, and S.~Schaal,
  ``Dynamical movement primitives: learning attractor models for motor
  behaviors,'' \emph{Neural computation}, vol.~25, no.~2, pp. 328--373, 2013.

\bibitem{hazara2019transferring}
M.~Hazara and V.~Kyrki, ``Transferring generalizable motor primitives from
  simulation to real world,'' \emph{IEEE Robotics and Automation Letters},
  vol.~4, no.~2, pp. 2172--2179, 2019.

\bibitem{haarnoja2018composable}
T.~Haarnoja, V.~Pong, A.~Zhou, M.~Dalal, P.~Abbeel, and S.~Levine, ``Composable
  deep reinforcement learning for robotic manipulation,'' in \emph{2018 IEEE
  International Conference on Robotics and Automation (ICRA)}.\hskip 1em plus
  0.5em minus 0.4em\relax IEEE, 2018, pp. 6244--6251.

\bibitem{lippi2020latent}
M.~Lippi, P.~Poklukar, M.~C. Welle, A.~Varava, H.~Yin, A.~Marino, and
  D.~Kragic, ``Latent space roadmap for visual action planning of deformable
  and rigid object manipulation,'' \emph{arXiv preprint arXiv:2003.08974},
  2020.

\bibitem{gothoskar2020learning}
N.~Gothoskar, M.~L{\'a}zaro-Gredilla, A.~Agarwal, Y.~Bekiroglu, and D.~George,
  ``Learning a generative model for robot control using visual feedback,''
  \emph{arXiv preprint arXiv:2003.04474}, 2020.

\bibitem{igl2018deep}
M.~Igl, L.~Zintgraf, T.~A. Le, F.~Wood, and S.~Whiteson, ``Deep variational
  reinforcement learning for pomdps,'' \emph{arXiv preprint arXiv:1806.02426},
  2018.

\bibitem{buesing2018learning}
L.~Buesing, T.~Weber, S.~Racaniere, S.~Eslami, D.~Rezende, D.~P. Reichert,
  F.~Viola, F.~Besse, K.~Gregor, D.~Hassabis \emph{et~al.}, ``Learning and
  querying fast generative models for reinforcement learning,'' \emph{arXiv
  preprint arXiv:1802.03006}, 2018.

\bibitem{mishra2017prediction}
N.~Mishra, P.~Abbeel, and I.~Mordatch, ``Prediction and control with temporal
  segment models,'' in \emph{Proceedings of the 34th International Conference
  on Machine Learning-Volume 70}.\hskip 1em plus 0.5em minus 0.4em\relax JMLR.
  org, 2017, pp. 2459--2468.

\bibitem{ke2018modeling}
N.~R. Ke, A.~Singh, A.~Touati, A.~Goyal, Y.~Bengio, D.~Parikh, and D.~Batra,
  ``Modeling the long term future in model-based reinforcement learning,''
  2018.

\bibitem{hafner2018learning}
D.~Hafner, T.~Lillicrap, I.~Fischer, R.~Villegas, D.~Ha, H.~Lee, and
  J.~Davidson, ``Learning latent dynamics for planning from pixels,''
  \emph{arXiv preprint arXiv:1811.04551}, 2018.

\bibitem{rhinehart2018deep}
N.~Rhinehart, R.~McAllister, and S.~Levine, ``Deep imitative models for
  flexible inference, planning, and control,'' \emph{arXiv preprint
  arXiv:1810.06544}, 2018.

\bibitem{krupnik2019multi}
O.~Krupnik, I.~Mordatch, and A.~Tamar, ``Multi agent reinforcement learning
  with multi-step generative models,'' \emph{arXiv preprint arXiv:1901.10251},
  2019.

\bibitem{brock2018large}
A.~Brock, J.~Donahue, and K.~Simonyan, ``Large scale gan training for high
  fidelity natural image synthesis,'' \emph{arXiv preprint arXiv:1809.11096},
  2018.

\bibitem{wang2018high}
T.-C. Wang, M.-Y. Liu, J.-Y. Zhu, A.~Tao, J.~Kautz, and B.~Catanzaro,
  ``High-resolution image synthesis and semantic manipulation with conditional
  gans,'' in \emph{Proceedings of the IEEE conference on computer vision and
  pattern recognition}, 2018, pp. 8798--8807.

\bibitem{vae_anom}
F.~{Wiewel} and B.~{Yang}, ``Continual learning for anomaly detection with
  variational autoencoder,'' in \emph{ICASSP 2019 - 2019 IEEE International
  Conference on Acoustics, Speech and Signal Processing (ICASSP)}, 2019, pp.
  3837--3841.

\bibitem{vae_text8672806}
W.~{Xu} and Y.~{Tan}, ``Semisupervised text classification by variational
  autoencoder,'' \emph{IEEE Transactions on Neural Networks and Learning
  Systems}, vol.~31, no.~1, pp. 295--308, 2020.

\bibitem{challenging_common}
F.~Locatello, S.~Bauer, M.~Lučić, G.~Rätsch, S.~Gelly, B.~Schölkopf, and
  O.~F. Bachem, ``Challenging common assumptions in the unsupervised learning
  of disentangled representations,'' in \emph{International Conference on
  Machine Learning}, 2019, best Paper Award.

\bibitem{IS_NIPS2016_6125}
T.~Salimans, I.~Goodfellow, W.~Zaremba, V.~Cheung, A.~Radford, X.~Chen, and
  X.~Chen, ``Improved techniques for training gans,'' in \emph{Advances in
  Neural Information Processing Systems 29}, D.~D. Lee, M.~Sugiyama, U.~V.
  Luxburg, I.~Guyon, and R.~Garnett, Eds.\hskip 1em plus 0.5em minus
  0.4em\relax Curran Associates, Inc., 2016, pp. 2234--2242.

\bibitem{FID_NIPS2017_7240}
M.~Heusel, H.~Ramsauer, T.~Unterthiner, B.~Nessler, and S.~Hochreiter, ``Gans
  trained by a two time-scale update rule converge to a local nash
  equilibrium,'' in \emph{Advances in Neural Information Processing Systems
  30}.\hskip 1em plus 0.5em minus 0.4em\relax Curran Associates, Inc., 2017,
  pp. 6626--6637.

\bibitem{binkowski2018demystifying}
M.~Bińkowski, D.~J. Sutherland, M.~Arbel, and A.~Gretton, ``Demystifying {MMD}
  {GAN}s,'' in \emph{International Conference on Learning Representations},
  2018.

\bibitem{sajjadi2018assessing}
M.~S. Sajjadi, O.~Bachem, M.~Lucic, O.~Bousquet, and S.~Gelly, ``Assessing
  generative models via precision and recall,'' in \emph{Advances in Neural
  Information Processing Systems}, 2018, pp. 5228--5237.

\bibitem{revisiting_pr}
L.~Simon, R.~Webster, and J.~Rabin, ``Revisiting precision recall definition
  for generative modeling,'' in \emph{Proceedings of the 36th International
  Conference on Machine Learning}, ser. Proceedings of Machine Learning
  Research, K.~Chaudhuri and R.~Salakhutdinov, Eds., vol.~97.\hskip 1em plus
  0.5em minus 0.4em\relax Long Beach, California, USA: PMLR, 09--15 Jun 2019,
  pp. 5799--5808.

\bibitem{higgins2018towards}
I.~Higgins, D.~Amos, D.~Pfau, S.~Racaniere, L.~Matthey, D.~Rezende, and
  A.~Lerchner, ``Towards a definition of disentangled representations,''
  \emph{arXiv preprint arXiv:1812.02230}, 2018.

\bibitem{repr_learning_survey}
Y.~{Bengio}, A.~{Courville}, and P.~{Vincent}, ``Representation learning: A
  review and new perspectives,'' \emph{IEEE Transactions on Pattern Analysis
  and Machine Intelligence}, vol.~35, no.~8, pp. 1798--1828, 2013.

\bibitem{tschannen2018recent}
M.~Tschannen, O.~Bachem, and M.~Lucic, ``Recent advances in autoencoder-based
  representation learning,'' \emph{arXiv preprint arXiv:1812.05069}, 2018.

\bibitem{kim2018disentangling}
H.~Kim and A.~Mnih, ``Disentangling by factorising,'' \emph{arXiv preprint
  arXiv:1802.05983}, 2018.

\bibitem{eastwood2018framework}
C.~Eastwood and C.~K. Williams, ``A framework for the quantitative evaluation
  of disentangled representations,'' 2018.

\bibitem{chen2018isolating}
T.~Q. Chen, X.~Li, R.~B. Grosse, and D.~K. Duvenaud, ``Isolating sources of
  disentanglement in variational autoencoders,'' in \emph{Advances in Neural
  Information Processing Systems}, 2018, pp. 2610--2620.

\bibitem{kumar2017variational}
A.~Kumar, P.~Sattigeri, and A.~Balakrishnan, ``Variational inference of
  disentangled latent concepts from unlabeled observations,'' \emph{arXiv
  preprint arXiv:1711.00848}, 2017.

\bibitem{jeon2019ibgan}
I.~Jeon, W.~Lee, and G.~Kim, ``{IB}-{GAN}: Disentangled representation learning
  with information bottleneck {GAN},'' 2019.

\bibitem{lee2020high}
W.~Lee, D.~Kim, S.~Hong, and H.~Lee, ``High-fidelity synthesis with
  disentangled representation,'' \emph{arXiv preprint arXiv:2001.04296}, 2020.

\bibitem{liu2019oogan}
B.~Liu, Y.~Zhu, Z.~Fu, G.~de~Melo, and A.~Elgammal, ``Oogan: Disentangling gan
  with one-hot sampling and orthogonal regularization,'' 2019.

\bibitem{kingma2014auto}
D.~P. Kingma and M.~Welling, ``Auto-encoding variational bayes,''
  \emph{International Conference on Learning Representations}, 2014.

\bibitem{rezende2014stochasticvae2}
D.~J. Rezende, S.~Mohamed, and D.~Wierstra, ``Stochastic backpropagation and
  approximate inference in deep generative models,'' in \emph{Int. Conf. Mach.
  Learn.}, 2014, pp. 1278--1286.

\bibitem{goodfellow2014generative}
I.~Goodfellow, J.~Pouget-Abadie, M.~Mirza, B.~Xu, D.~Warde-Farley, S.~Ozair,
  A.~Courville, and Y.~Bengio, ``Generative adversarial nets,'' in
  \emph{Advances in neural information processing systems}, 2014, pp.
  2672--2680.

\bibitem{disentanglement_video}
\BIBentryALTinterwordspacing
E.~L. Denton and v.~Birodkar, ``Unsupervised learning of disentangled
  representations from video,'' in \emph{Advances in Neural Information
  Processing Systems 30}, I.~Guyon, U.~V. Luxburg, S.~Bengio, H.~Wallach,
  R.~Fergus, S.~Vishwanathan, and R.~Garnett, Eds.\hskip 1em plus 0.5em minus
  0.4em\relax Curran Associates, Inc., 2017, pp. 4414--4423. [Online].
  Available:
  \url{http://papers.nips.cc/paper/7028-unsupervised-learning-of-disentangled-representations-from-video.pdf}
\BIBentrySTDinterwordspacing

\bibitem{creager2019flexibly}
E.~Creager, D.~Madras, J.-H. Jacobsen, M.~A. Weis, K.~Swersky, T.~Pitassi, and
  R.~Zemel, ``Flexibly fair representation learning by disentanglement,''
  \emph{arXiv preprint arXiv:1906.02589}, 2019.

\bibitem{lee2018diverse}
H.-Y. Lee, H.-Y. Tseng, J.-B. Huang, M.~Singh, and M.-H. Yang, ``Diverse
  image-to-image translation via disentangled representations,'' in
  \emph{Proceedings of the European conference on computer vision (ECCV)},
  2018, pp. 35--51.

\bibitem{van2019disentangled}
S.~van Steenkiste, F.~Locatello, J.~Schmidhuber, and O.~Bachem, ``Are
  disentangled representations helpful for abstract visual reasoning?''
  \emph{arXiv}, pp. arXiv--1905, 2019.

\bibitem{JMLR:v13:gretton12a}
A.~Gretton, K.~M. Borgwardt, M.~J. Rasch, B.~Sch{{\"o}}lkopf, and A.~Smola, ``A
  kernel two-sample test,'' \emph{Journal of Machine Learning Research},
  vol.~13, no.~25, pp. 723--773, 2012.

\bibitem{coleman2014reducing}
D.~Coleman, I.~Sucan, S.~Chitta, and N.~Correll, ``Reducing the barrier to
  entry of complex robotic software: a moveit! case study,'' \emph{arXiv
  preprint arXiv:1404.3785}, 2014.

\end{thebibliography}
